\documentclass[10pt,twocolumn,letterpaper]{article}

\usepackage{iccv}
\usepackage{times}
\usepackage{epsfig}
\usepackage{graphicx}
\usepackage{amsmath}
\usepackage{amssymb}
\usepackage{multirow}
\usepackage{enumitem}
\usepackage[pagebackref=true,breaklinks=true,letterpaper=true,colorlinks,bookmarks=false]{hyperref}

\iccvfinalcopy % *** Uncomment this line for the final submission

\def\iccvPaperID{1134} % *** Enter the ICCV Paper ID here

\ificcvfinal\fi

\begin{document}

%%%%%%%%% TITLE
\title{BSS-Bench: Towards Reproducible and Effective Band Selection Search}

\author{Wenshuai Xu\\
Beihang University\\
\and
Zhenbo Xu\\
Beihang University\\
}

\maketitle
\ificcvfinal\thispagestyle{empty}\fi

%%%%%%%%% ABSTRACT
\begin{abstract}
   %Hyperspectral analysis has tremendous potential in various fields due to the rich information in hyperspectral data. 
   The key technology to overcome the drawbacks of hyperspectral imaging (expensive, high capture delay, and low spatial resolution) and make it widely applicable is to select only a few representative bands from hundreds of bands. However, current band selection (BS) methods face challenges in fair comparisons due to inconsistent train/validation settings, including the number of bands, dataset splits, and retraining settings. To make BS methods easy and reproducible, this paper presents the first band selection search benchmark (BSS-Bench) containing $52$k training and evaluation records of numerous band combinations (BC) with different backbones for various hyperspectral analysis tasks. The creation of BSS-Bench required a significant computational effort of $1.26$k GPU days. By querying BSS-Bench, BS experiments can be performed easily and reproducibly, and the gap between the searched result and the best achievable performance can be measured. Based on BSS-Bench, we further discuss the impact of various factors on BS, such as the number of bands, unsupervised statistics, and different backbones. In addition to BSS-Bench, we present an effective one-shot BS method called Single Combination One Shot (SCOS), which learns the priority of any BCs through one-time training, eliminating the need for repetitive retraining on different BCs. Furthermore, the search process of SCOS is flexible and does not require training, making it efficient and effective. Our extensive evaluations demonstrate that SCOS outperforms current BS methods on multiple tasks, even with much fewer bands. Our BSS-Bench and codes are available in the supplementary material and will be publicly available.
\end{abstract}
%Our analysis on BSS-Bench result in three insights about BS. By effective BS, three bands are enough to achieve high performances comparable to the state-of-the-art methods using 30 bands. Supervised learning-based BS method has the potential to select better BCs than unsupervised approaches. The superiority of different BCs is less affected by different backbones.

\begin{figure}[!t]
\centering
\includegraphics[width=1.0\linewidth]{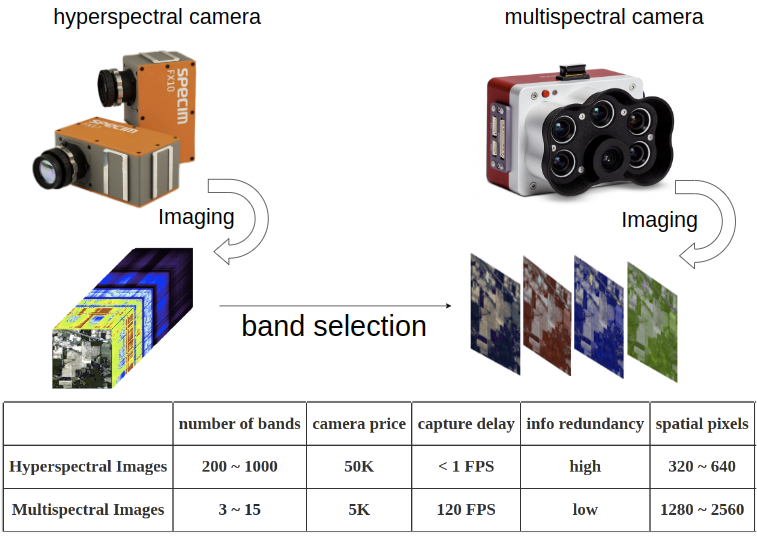}
\caption{Relationship between HS imaging, MS imaging, and band selection. MS cameras can achieve high analysis performances similar to HS cameras by exploiting only a few representative bands found by band selection algorithms, while at the same time enjoying many advantages such as lower capture latency, affordability, and higher spatial resolutions.
}
\label{intro_bs}
\end{figure}

%%%%%%%%% TODO: 为什么是NBS而不是BS
\section{Introduction}
Deep learning-based techniques, such as multi-modal large language models, have made remarkable progress in processing RGB images \cite{driess2023palme,ramesh2022hierarchical}. However, the perception based solely on RGB cameras is limited. As the ultimate physical footprint of any material \cite{li2023jointly,stuart2022high}, hyperspectral image (HSI) analysis has demonstrated enormous potential in various fields such as remote sensing \cite{grewal2023machine}, precision agriculture \cite{yang2023double}, environmental monitoring \cite{alkhatib2023tri}, bioanalytical analyses, food safety and quality analysis, pharmaceutical tablet manufacturing \cite{wang2023simple}. 
Despite being powerful, hyperspectral (HS) cameras have several well-known drawbacks, such as being costly, having high capture latency, and low spatial resolution.
On the other hand, as shown in Fig. \ref{intro_bs}, multispectral (MS) imaging, which utilizes a limited number of spectral bands, can perform with high throughput, high resolution, high robustness, competitive pricing, and ease of use.
Therefore, Band selection (BS) \cite{tang2021hyperspectral} is the key technology for extending the capabilities of HS imaging to a broader range of applications. Unlike dimensionality reduction techniques \cite{rezasoltani2023hyperspectral,yang2023double} in HSI analysis that project high-dimensional HSI features onto fewer channels, BS algorithms aim to select a few bands (usually $3\sim15$) from hundreds of bands in HS imaging.
Then, CMOS technologies can then be employed to create a custom filter mosaic for building an MS camera that captures images consisting of the selected spectral bands \cite{meng2021self}.
In general, the effectiveness of BS has a significant impact on the performance of MS cameras.

Despite significant progress in recent BS methods \cite{li2023jointly,li2023hyperspectral,yang2023double,morales2021hyperspectral,sun2021multiscale}, it remains challenging to make fair comparisons between different methods due to inconsistent train/validation settings, including the number of bands, dataset splits, and retraining settings.
%the use of different datasets, the number of selected bands, and training or fine-tuning settings.
Moreover, the optimal band selection choice is unknown, making the gap between the search result and the optimal result unclear to the community. 
Furthermore, the search for the best band combinations (BCs) requires vast computational resources, which poses a significant barrier to the advancement of BS.
We find that the development of BS is similar to the development of NAS. 
Several years ago, NAS research \cite{mills2022aio,xiao2022shapley,ye2023beta} were criticized for producing non-reproducible results and performance sensitivity to specific training pipelines, hyper-parameters, and even random seeds.
In response, NAS benchmarks \cite{dong2020bench,ying2019bench,zela2022surrogate} were introduced to address these issues. These benchmarks propose a compact architecture dataset containing hundreds of thousands of unique architectures and their evaluation results, which helps ameliorate the difficulty of reproduction and relieve the burden of large-scale computations.
%Although great success has been made, most BS methods \cite{li2023jointly,li2023hyperspectral,yang2023double,morales2021hyperspectral,sun2021multiscale} validate on different datasets with different number of selected bands and different training or fine-tuning settings, making the fair comparisons of different BS methods challenging. 
%Lastly, due to the demand of large computational resources, the search of tremendous band combinations (BCs) becomes a significant barrier to the advancement of BS.
%NAS researches \cite{mills2022aio,xiao2022shapley,ye2023beta} had been criticized substantially three years ago for non-reproducible works, strong sensitivity of performances to carefully-chosen training pipelines, hyper-parameters and even random seeds \cite{ying2019bench}. 
%To handle these issues, NAS benchmarks \cite{dong2020bench,ying2019bench,zela2022surrogate} had been proposed in recent years to ameliorate the difficulty of reproduction and relief the burden of large-scale computations by proposing a compact architecture dataset containing hundreds of thousands of unique architectures and their evaluation results. 
%Inspired by the rapid development of neural architecture search (NAS) that processes the search space of neural architectures, we think that 
%We think that the development of BS can draw inspirations from the advancement of NAS.

Inspired by the advancements of NAS, this paper presents the first Band Selection Search Benchmark (BSS-Bench) to make BS research easy and reproducible. We trained and evaluated a huge number of different BCs with two different backbones (CNN and Transformer) for two different tasks (reconstruction \cite{cai2022mst++} and classification \cite{alkhatib2023tri}), utilizing over 1.26k GPU days of computation time. The results are compiled into a large table that maps $52$k BCs with varying backbones and tasks to task-specific metrics. 
BSS-Bench enables BS algorithms to be performed by querying the benchmark table rather than performing the actual training and evaluation process.
Moreover, as the training settings are the same for all BCs, comparisons between different BS algorithms can be made fairly, and the deviation from the optimal BS result can also be measured.

Alongside BSS-Bench, we present an effective searching technique for one-shot BS named Single Combination One Shot (SCOS) that jointly learns the priority of any BCs by designing a novel spectral spatial position embedding (SSPE). We compare three different position embeddings in Table \ref{ablation_study} and find that the separately learnable SSPE achieves the best performance. 
Once trained, the SCOS model can evaluate the performance of any BCs without the need for re-training. The BC with the highest evaluation result is considered the search result.

%In evaluations, We have implemented various BS methods and validate them on the BSS-Bench setting. Comprehensive experiments verify that our method SCOS achieves superior BS results over current methods on different tasks. 
%The search cost is significantly less than other learning based methods \cite{wen2020neural,ye2022band}. 
%In addition, we perform systematic analysis on the impact of the number of bands, the unsupervised statistics, and different backbones, yielding several insights on BS.
We conduct extensive evaluations of various BS methods \cite{morales2021hyperspectral,mou2021deep} on BSS-Bench, and our SCOS outperforms current methods on different tasks (see Table \ref{main_result}).
Moreover, our method incurs significantly lower search costs than other learning-based methods \cite{wen2020neural,ye2022band}.
Through systematic analysis of the impact of the number of bands, unsupervised statistics, and different backbones, we gained several insights into BS.  
Firstly, with effective searching algorithms, three bands are enough to achieve high performance, which is comparable to state-of-the-art methods that use hundreds of bands.
Secondly, supervised learning-based BS methods have the potential to select better BCs than unsupervised approaches. 
Thirdly, the superiority of different BCs is less affected by different backbones.

Our main contributions are summarized as follows:
%\begin{itemize}[noitemsep,nolistsep,topsep=0pt]
%Representative BS methods have been implemented on BSS-Bench. 
\begin{itemize}[noitemsep,nolistsep]
    \item We construct the first BS Benchmark named BSS-Bench to make research easy and reproducible by exhaustively evaluating $52$k unique BCs on different tasks and backbones.
    \item An effective BS method named Single Combination One Shot (SCOS) is proposed to learn the priority of any band combinations by one-time training. The search process is training-free, fast, and flexible.
    \item We carry out systematic analysis on BSS-Bench to provide various insights into BS. Extensive evaluations show that our SCOS achieves superior performance over current methods, even with much fewer bands.
\end{itemize}

\section{Related Work}
%supervised methods, refer to <ye2022band>
\textbf{Band Selection}
Recent BS approaches can be grouped into three types: unsupervised BS algorithms \cite{mou2021deep,su2018saliency}, semi-supervised BS methods, and supervised BS methods. Most unsupervised BS methods select spectral bands by clustering \cite{morales2021hyperspectral,sun2021multiscale}, ranking \cite{sarhrouni2012band,xu2021similarity}, or searching \cite{li2023hyperspectral,mou2021deep}. Different from unsupervised BS methods, semi-supervised approaches \cite{feng2021deep} usually exploit both unlabeled and labeled samples for learning the weight of bands. Current supervised BS methods \cite{yang2023double,ye2022band} include two categories: sequential search \cite{li2021improved}, and stochastic search \cite{ding2020improved,ye2022band}. Sequential search methods usually consist of forward search, backward search, and floating forward search \cite{li2021improved}. Recent works have proposed a number of nature-inspired stochastic search algorithms for BS, such as genetic algorithm, gray wolf optimizer \cite{al2020review}, and hybrid rice optimization \cite{ye2022band}. 
However, to the best of our knowledge, there are few works that perform exhaustive search on hyperspectral image datasets. Our BSS-Bench is the first BS benchmark that explores the search space of BS to facilitate reproducible research. In addition to BSS-Bench, we propose a novel one-shot learning-based BS method named SCOS.
Our SCOS designs effective spectral spatial encoding ways to enable joint learning of any BCs and the network weights. Evaluations on Table \ref{main_result} show that SCOS achieves better BS results than current methods.
%The genetic algorithm MHRO proposes an opposition-based-learning strategy and differential evolution operators to increase the diversity of the population.

\begin{figure}[!t]
\centering
\includegraphics[width=0.49\linewidth]{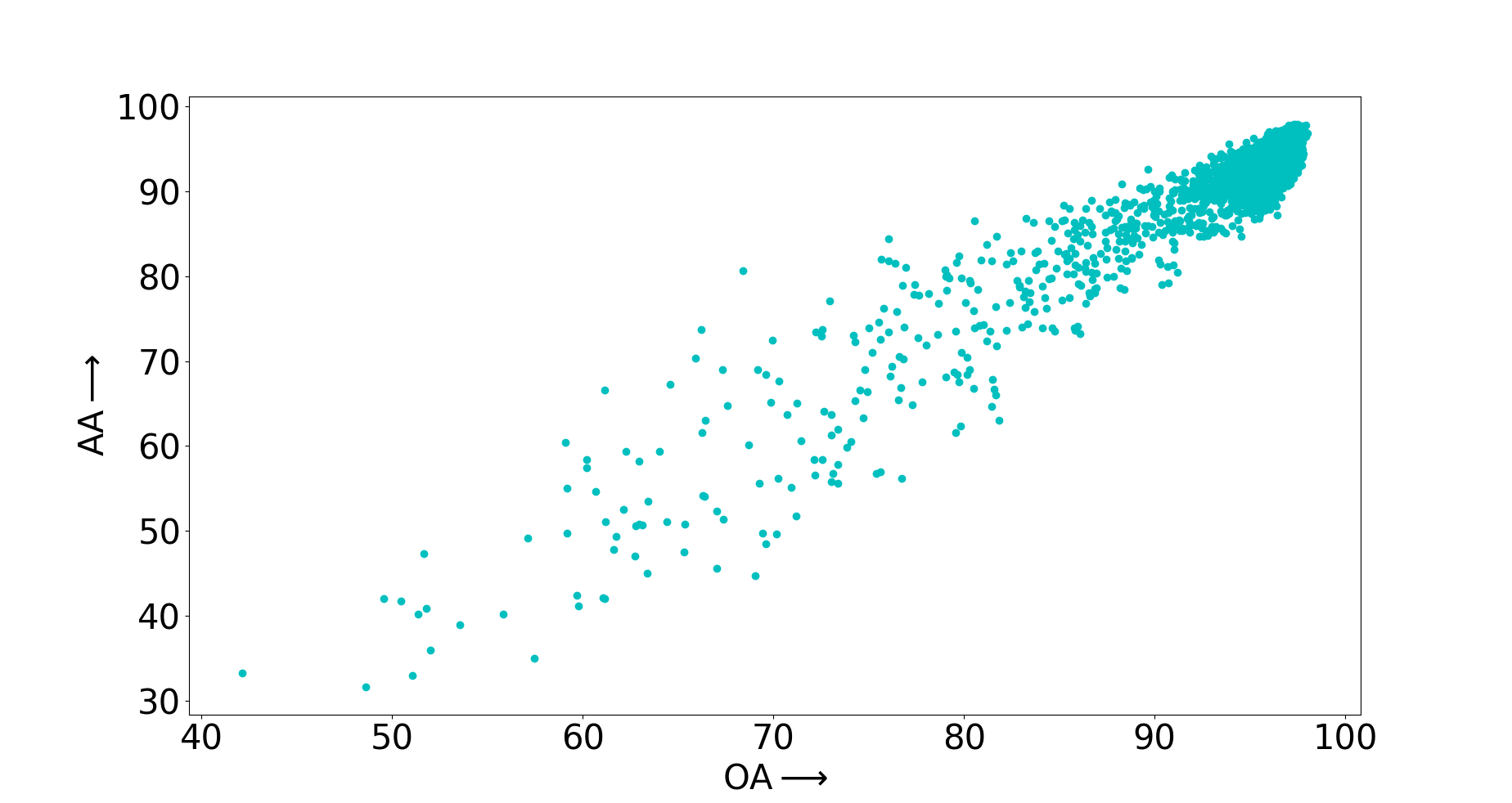}
\includegraphics[width=0.49\linewidth]{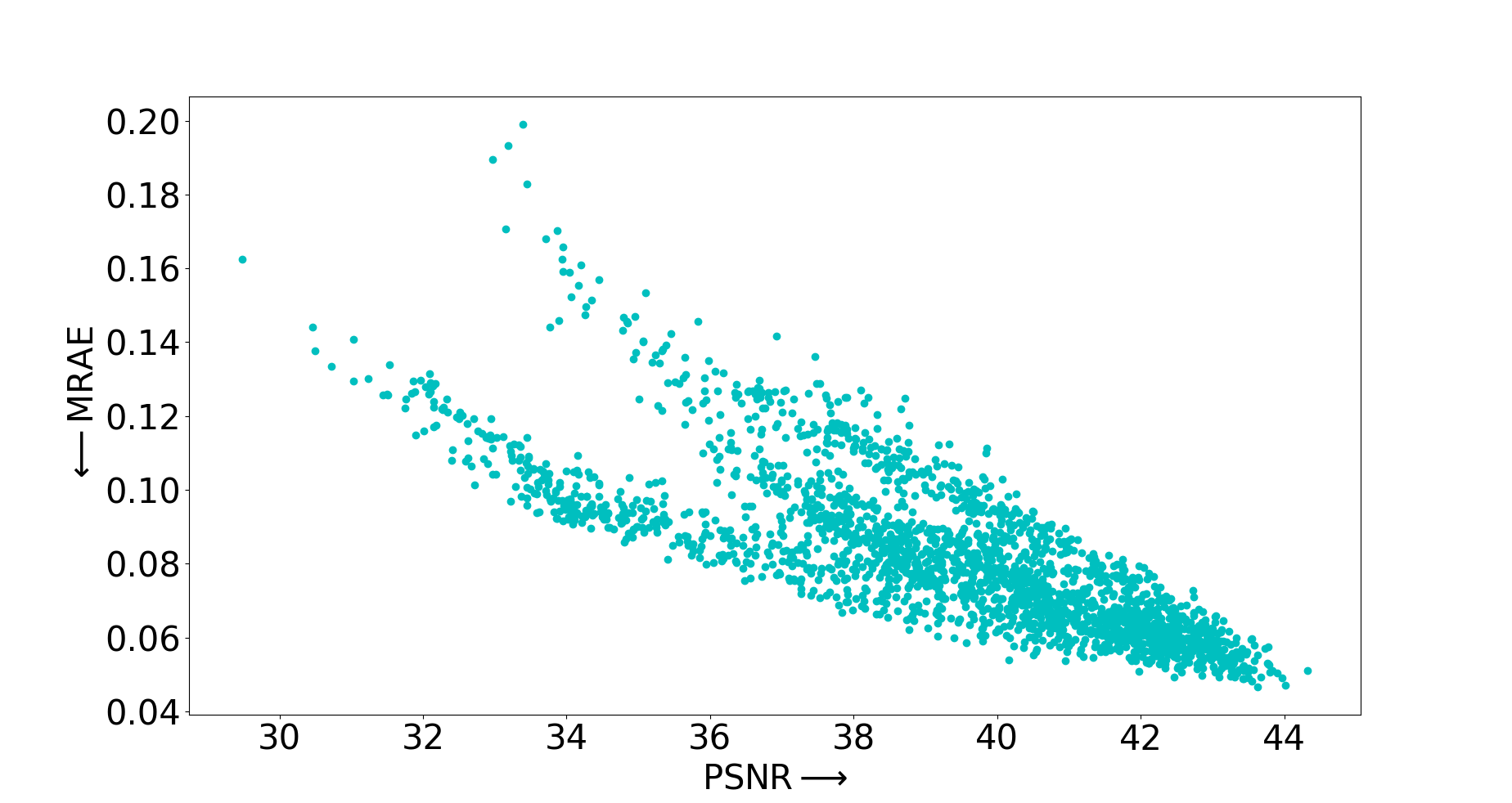}
\caption{The scatter plot of different band combinations on the HSI classification task (left) and the HSI reconstruction task (right). The performances of various band combinations vary significantly.
Best viewed in color.
}
\label{intro_scatter}
\end{figure}

\textbf{Hyperspectral image classification}
Recent HSI classification research adopt conventional machine learning techniques such as support vector machines, dimension reduction \cite{sun2022spectral}, kernel-based algorithms \cite{grewal2023machine}, and deep learning-based approaches \cite{alkhatib2023tri,yang2023double}. 
Support vector machine-based methods \cite{rani2022machine} learn an optimal hyperplane that separates high-dimensional hyperspectral features well.
Dimension reduction approaches \cite{grewal2023machine} solve the dimensionality curse of HSI data by selecting the most important characteristics using principal component analysis or independent component analysis.
Kernel-based algorithms \cite{gu2017multiple} formulate the non-linear and complex nature of HSIs as the kernel consisting of mathematical functions.
Many recent works \cite{cai2022mst++} focus on incorporating deep learning techniques to extract features automatically, solve non-linearity problems and achieve high performances. 
Our BSS-Bench incorporates the Indian Pines dataset for the HSI classification task. Fig. \ref{intro_scatter} shows that the performances of different BCs vary significantly. Moreover, 
BSS-Bench shows that three carefully selected bands have the potential to achieve comparable HSI classification performance to current methods using 200 bands (see Table \ref{ip_compare}).

\textbf{Hyperspectral image reconstruction}
% TODO: add denoise paper
HSI reconstruction aims to restore HSIs from raw measurements \cite{dauhst}. Model-based methods \cite{liu2018rank} rely on manually designed image priors like variation or sparsity and parameter tweaking. Plug-and-play algorithms \cite{meng2021self,qiao2020deep,yuan2020plug,yuan2021plug} usually adapt pre-trained image restoration networks into conventional model-based methods to achieve HSI reconstruction.
End-to-end (E2E) approaches \cite{hu2022hdnet,miao2019net} typically deploy convolutional neural networks (CNN) to learn a mapping function from a measurement to HSIs in an end-to-end manner.
Many recent works propose deep unfolding methods \cite{huang2021deep,zhang2022herosnet} to design a multi-stage architecture to reconstruct the HSI cube based on the corresponding measurement by explicitly characterizing the imaging model and image priors. In our BSS-Bench, both CNN and transformer architectures are tested on HSI reconstruction. Moreover, our SCOS can search for effective BCs that approach the best achievable performance.

\section{The BSS-Bench}\label{bench_section}
Our BSS-Bench is a table that maps numerous BCs with different backbones and spectral analysis tasks to their training and evaluation metrics. An overview of BSS-Bench is available in Table \ref{bench_intro_table}.
To date, we have constructed datasets for the following tasks: HSI reconstruction \cite{cai2022mst++} and HSI classification \cite{alkhatib2023tri}. 
Moreover, without loss of generality, we selected two kinds of backbones: CNN and transformer. 
Comparisons between the two backbones with respect to the same BCs on different tasks are available in Fig. \ref{different_architecture_impact}. 
Note that we will continue to enrich the number of data sets and the number of tasks to promote the advancement of more spectral analysis tasks.
Based on our BSS-Bench, unnecessary repetitive training procedures of each selected BC can be eliminated so that researchers can focus on the essence of BS, i.e., searching for the best BC. Another advantage is that the validation cost for BS largely decreases when testing in our BSS-Bench, which provides a computationally friendly environment for more participations in BS. Hopefully, BSS-Bench will show its value in the field of spectral band selection research.
The construction of the BSS-Bench consists of the following four steps. It is worth noting that both our BSS-Bench and codes are available in the supplementary material.
% Unnecessary repetitive training procedure of each selected architecture can be avoided (Liu et al., 2018; Zoph & Le, 2017) so that researchers can target on the essence of NAS, i.e., search algorithm. Another benefit is that the validation time for NAS largely decreases when testing in NAS-Bench-201, which provides a computational power friendly environment for more participations in NAS.

\begin{table}[!t]
\centering
\resizebox{1.0\linewidth}{!}{%
\begin{tabular}{|cl|cc|ccc|}
\hline
\multicolumn{2}{|l|}{Tasks}                                        & \multicolumn{2}{c|}{HSI reconstruction} & \multicolumn{3}{c|}{HSI classification}                      \\ \hline
\multicolumn{2}{|l|}{Datasets}                                     & \multicolumn{2}{c|}{NTIRE 2022}         & \multicolumn{3}{c|}{Indian Pines}                            \\ \hline
\multicolumn{2}{|l|}{Metrics}                                      & \multicolumn{1}{c|}{MRAE}     & PSNR    & \multicolumn{1}{c|}{OA} & \multicolumn{1}{c|}{AA} & Kappa    \\ \hline
\multicolumn{2}{|l|}{Architectures}                                & \multicolumn{1}{c|}{HINET}    & MST++   & \multicolumn{2}{c|}{SSRN}                         & SSFTTnet \\ \hline
\multicolumn{1}{|c|}{\multirow{6}{*}{HP.}} & Iterations    & \multicolumn{2}{c|}{50K}                & \multicolumn{3}{c|}{16K}                                     \\ \cline{2-7} 
\multicolumn{1}{|c|}{}                            & Patch Size    & \multicolumn{2}{c|}{32x32}              & \multicolumn{3}{c|}{13x13}                                   \\ \cline{2-7} 
\multicolumn{1}{|c|}{}                            & Batch Size    & \multicolumn{2}{c|}{128}                & \multicolumn{3}{c|}{64}                                      \\ \cline{2-7} 
\multicolumn{1}{|c|}{}                            & Learning Rate & \multicolumn{2}{c|}{4e-4}               & \multicolumn{3}{c|}{1e-3}                                    \\ \cline{2-7} 
\multicolumn{1}{|c|}{}                            & LR scheduler  & \multicolumn{2}{c|}{CosineAnnealingLR}  & \multicolumn{3}{c|}{CosineAnnealingLR}                       \\ \cline{2-7} 
\multicolumn{1}{|c|}{}                            & Optimizer     & \multicolumn{2}{c|}{AdamW}              & \multicolumn{3}{c|}{Adam}                                    \\ \hline
\multicolumn{2}{|l|}{NBC}                       & \multicolumn{2}{c|}{4495}               & \multicolumn{3}{c|}{21600}                                    \\ \hline
\multicolumn{2}{|l|}{Cost (GPU days)}                                & \multicolumn{1}{c|}{206.02}    & 480.76   & \multicolumn{2}{c|}{282.72}                         & 294.86 \\ \hline
\end{tabular}
}
\caption{The characteristics of BSS-Bench. HP. represents hyper-parameters. NBC denotes the number of band combinations.}
\label{bench_intro_table}
\end{table}

\textbf{Step 1: select tasks, datasets, and metrics.}
Existing works on hyperspectral analysis mainly focus on HSI classification and HSI reconstruction \cite{hu2022hdnet} due to the limited types of HSI datasets \cite{grewal2023machine}. Therefore, we consider these two mainstream tasks in BSS-Bench: HSI classification and HSI reconstruction. 
For classification, we choose the popular Indian Pines Dataset \cite{baumgardner2015220}, which has $200$-bands HSI data for land cover classification. The overall accuracy (OA), the average accuracy (AA), and the Kappa coefficient are selected as the evaluation metrics following previous works \cite{sun2022spectral}.
For reconstruction, we adopt the recent NTIRE 2022 dataset \cite{arad2022ntire}, which has $31$-bands HSI data for reconstruction. The mean relative absolute error (MRAE) and the Peak Signal-to-Noise Ratio (PSNR) are computed in the HSI classification experiments.
%Note that we will continual to include more HSI classification datasets following similar procedures introduced in this section.
%Few pioneering works \cite{yan2021object} consider to solve the object detection problem based on current HSI datasets.

\textbf{Step 2: select backbones and search hyper-parameters.}
Without loss of generality, we compare two types of backbones: CNN and transformer. 
For CNN, we adopt the effective HINet \cite{chen2021hinet} due to its effectiveness in HSI reconstruction \cite{cai2022mst++} and the SSRN \cite{zhong2017spectral} due to its good performance in HSI classification. For Transformer, we exploit the recent transformer-based architecture MST++ \cite{cai2022mst++} for HSI reconstruction and SSFTT \cite{sun2022spectral} for HSI classification. 
To train each backbone on each dataset, we carry out hyper-parameter optimization search \cite{Microsoft_Neural_Network_Intelligence_2021} based on perturbations of their original training settings.
The searched training hyper-parameters, such as patch size and learning rate for both HINet and SSFTT on these two tasks, are listed in Table \ref{bench_intro_table}.

\textbf{Step 3: band combination selection.}
To explore the best potential of BS, our BSS-Bench sets the number of bands to $3$ due to three reasons. 
Firstly, the redundancy of HSI information is very high as HSIs generally have hundreds of bands, resulting in the large capture delay, high complexity, and high price of optical components. Generally speaking, the fewer the number of bands to capture, the cheaper the MS camera. 
Secondly, three spectral bands have been proven effective in handling complex hyperspectral tasks like HSI reconstruction \cite{cai2022mst++}. In Table \ref{ip_compare}, the results also show that our method achieves state-of-the-art performances with only three spectral bands. 
Thirdly, methods that work well with a limited number of bands are expected to achieve better performance with an increase in the band count.
%We believe that the cost of MS camera with three bands can refer to the manufacturing process of existing RGB cameras.

For datasets with fewer spectral bands, such as the NTIRE 2022 dataset, we simply evaluate all possible BCs. The number of possible BCs is $\tbinom{31}{3}=4,495$. For HSI datasets with hundreds of spectral bands, we select a representative set of BCs in two steps. Let's take the Indian Pines dataset, for example, which has $200$ bands. The total number of possible BCs, denoted by $A$, is $\tbinom{200}{3}=1,313,400$. Searching through millions of choices is environmentally unfriendly, so we introduce an effective strategy called Predict-and-Expand. Firstly, we randomly select $5,000$ BCs as the initial set $I$ and evaluate the performances of these BCs to construct a training set for a lightweight accuracy predictor \cite{wen2020neural}. Then, we train the predictor and use the predictor to predict the performance of the remaining $1,308$k BCs. After that, the initial set is expanded by incorporating BCs with the top-performing predictions. For the Indian Pines dataset, the number of evaluated BCs reaches $21,600$.
% 1264.36 GPU days

\textbf{Step 4: exhaustive benchmarking.}
For each task, we train and evaluate all chosen BCs with both the CNN and transformer architectures. The number of BCs is $4,495$ for HSI reconstruction and $21,600$ for HSI classification. As we evaluate two different architectures exhaustively with all BCs, the total number of items in our BSS-Bench is $52,190$.
Each item represents the training and the evaluation of a CNN or Transformer backbone with a possible BC on HSI reconstruction or HSI classification. 
For each item, the model is trained twice using different random seeds while following the same training settings outlined in Table \ref{bench_intro_table}. 
The training logs and performance metrics for each item are provided, and performance on all evaluation metrics is accessible.
Moreover, the costs of GPU days for each combination of architectures and tasks are shown in the last row of Table \ref{bench_intro_table}. The construction of BSS-Bench requires a significant computation of $1.26$k GPU days in total.

\section{Method}\label{method_section}
In this section, we present a novel one-shot BS framework coined Single Combination One-Shot (SCOS). Our SCOS supports different kinds of spectral analysis tasks and is one-shot, meaning it only needs to be trained once. After training, the SCOS model can be exploited to predict the performance of any BCs. Then, the band combination with the top performance is regarded as the search result. 
As our framework focus on learning-based one-shot BS and is applicable to various spectral analysis tasks, we omit the detailed descriptions of two tasks (HSI reconstruction \cite{cai2022mst++} and HSI classification \cite{alkhatib2023tri}) and suggest that reviewers refer to the cited papers.

\begin{figure*}[!t]
\centering
\includegraphics[width=0.85\linewidth]{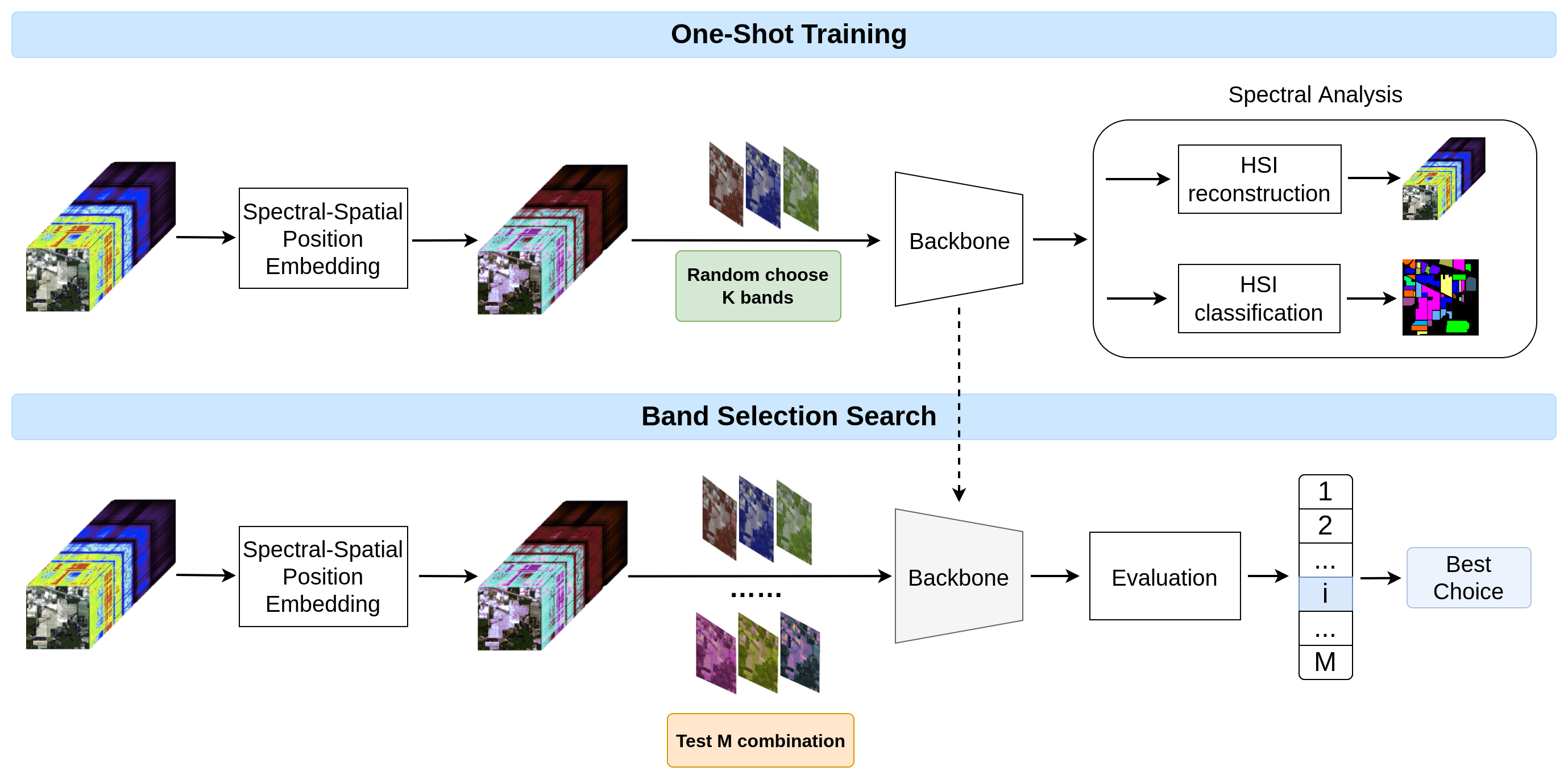}
\caption{The overview of our SCOS framework. We design a unified framework with effective spectral spatial position embeddings to learn inputs of arbitrary band combinations. After one-shot training, the backbone can evaluate the performance of any band combinations through pure inference.
The band combination with the top performance is chosen as the search result.
Best viewed in color.
}
\label{method}
\end{figure*}

Suppose that $A$ represents the search space of all BCs, a selected BC is one item $c_K \in A$.
The backbone processing $c_K$ is denoted as $\mathcal{N}(c_K, w)$ with weight $w$, where $K$ is the number of spectral bands to select.

Our learning-based method learns the priority of any BCs and the weights of the backbone in an end-to-end manner
and aims to solve two related problems. The first problem is to optimize weight $w$, 
\begin{equation}
    w_{c_K} = \underset{w}{\text{argmin}} \; \mathcal{L}_{\text{train}}(\mathcal{N}(c_K, w)),
\label{problem_1}
\end{equation}
where $\mathcal{L}_{\text{train}}$ is the training loss function of the corresponding task. 

The second problem is BS optimization, which searches for the BC that is trained on the training set and, at the same time, achieves the best evaluation result on validation metrics. 
The best BC ${c_K}^*$ can be obtained as follows:
\begin{equation}
    {c_K}^* = \underset{c_K \in A}{\text{argmax}} \; \text{Prec}_{\text{val}}(\mathcal{N}(c_K, w_{c_K}))
\label{problem_2}
\end{equation}
where $\text{Prec}_{\text{val}}(\cdot)$ represents the performance on the main validation metric. It is worth noting that evaluation in Eq. (\ref{problem_2}) only requires inference. Therefore, the search process is very efficient.

Most current BS methods solve the two optimization problems in a nested manner \cite{feng2021deep,li2021improved,li2023hyperspectral,morales2021hyperspectral,mou2021deep,sarhrouni2012band,xu2021similarity,ye2022band}, where countless BCs are chosen from $A$ and backbones are trained from scratch with each BC as formulated in Eq. (\ref{problem_1}). For HSI data with hundreds of bands, thousands of training sessions are usually required, resulting in environmentally unfriendly experiments and high computational demand. To relieve the burden of repeated training, SCOS decouples the backbone training and the band selection into two sequential steps, resulting in an efficient BS method that trains the backbone only once.

However, designing a unified framework to learn input with arbitrary BCs (often millions) is non-trivial. 
Directly feeding the three-band MS image as the traditional RGB images to the backbone eliminates the information of selected $K$ spectral wavelengths. Empirically we found that this makes the learning process difficult to converge.

Let $N$ represents the number of bands in HSI.
Given the input HSI patch image $I_H=\{p^{(u,v)}_i, i\in [1, N], u \in [1, W], v \in [1,H]\} \in  \mathbb{R}^{H\times W\times N}$, 
we transform $I_H$ into $\hat{I}_H=\{p^s_i, i\in [1, N], s \in [1,W*H]\} \in \mathbb{R}^{HW\times N}$ by merging two spatial dimensions.
$p^s_i$ represents the pixel at each spectral ($i$) and spatial ($s$) location. In order for the backbone to make use of the order of spectral wavelengths, we explore three different position embeddings as follows.

\textbf{Absolute PE (APE).} Inspired by the original position encoding in transformer, we adopt the sinusoidal function to encode the location of the spectral band as follows:
\begin{equation}
\begin{split}
    {E_{SS}}_{(i, 2j)} &= sin(i / 10000^{2j/(W*H)}) \\
    {E_{SS}}_{(i, 2j+1)} &= cos(i / 10000^{2j/(W*H)})
\end{split}
\label{absolute_pe_equ}
\end{equation}
where $j \in [1, (W*H)/2]$.

\textbf{Co-learnable PE (CLPE).} 
Following the learnable position embedding in ViT \cite{dosovitskiy2020image}, the position of spectral bands as well as the spatial locations are formulated as the co-learnable embedding $E_{SS} \in \mathbb{R}^{N\times HW}$.

For the absolute PE or the co-learnable PE, we add it to the transformed HSI input $\hat{I}_H$ and obtain the encoded HSI input $E_H \in \mathbb{R}^{N\times HW}$. $E_H$ is formulated as:
\begin{equation}
    E_H = \text{MLP}(\hat{I}_H) + E_{SS}
\end{equation}
where MLP denotes one-layer perception.

\textbf{Separately learnable PE (SLPE).} The expression of co-Learnable PE supposes that the embedding should be different for spatial pixels and spectral bands. Different from the co-learnable PE, we also introduce a separately learnable PE that learn two independent embeddings for spectral bands $E_b \in \mathbb{R}^{N\times 1}$ and spatial pixels $E_p \in \mathbb{R}^{1\times HW}$, respectively. In this way, the encoded HSI input $E_H$ is formulated as:
\begin{equation}
    E_H = \text{MLP}(\text{MLP}(\hat{I}_H) + E_p) + E_b
\end{equation}

\begin{table*}[!t]
\centering
\resizebox{0.7\linewidth}{!}{%
\begin{tabular}{|c|c|cccccc|}
\hline
\multirow{2}{*}{Task} & \multirow{2}{*}{Metrics} & \multicolumn{6}{c|}{Methods}                                                                                                                                                               \\ \cline{3-8} 
                      &                          & \multicolumn{1}{c|}{IBRA}         & \multicolumn{1}{c|}{Predictor} & \multicolumn{1}{c|}{MHRO}         & \multicolumn{1}{c|}{DRL}        & \multicolumn{1}{c|}{SCOS}         & Oracle      \\ \hline
                      \hline
\multirow{5}{*}{CLS.}  & Band Comb.           & \multicolumn{1}{c|}{61, 126, 163} & \multicolumn{1}{c|}{3, 6, 92}  & \multicolumn{1}{c|}{86, 117, 129} & \multicolumn{1}{c|}{34, 55, 64} & \multicolumn{1}{c|}{92, 121, 162} & 66, 92, 176 \\ \cline{2-8} 
                      & OA* (\%)                     & \multicolumn{1}{c|}{96.70}        & \multicolumn{1}{c|}{97.12}     & \multicolumn{1}{c|}{97.53}        & \multicolumn{1}{c|}{97.34}      & \multicolumn{1}{c|}{\textbf{97.64}}        & 98.01       \\ \cline{2-8} 
                      & AA (\%)                      & \multicolumn{1}{c|}{95.15}        & \multicolumn{1}{c|}{93.70}     & \multicolumn{1}{c|}{96.56}        & \multicolumn{1}{c|}{\textbf{97.48}}      & \multicolumn{1}{c|}{95.25}        & 96.87       \\ \cline{2-8} 
                      & Kappa                 & \multicolumn{1}{c|}{0.9624}        & \multicolumn{1}{c|}{0.9672}     & \multicolumn{1}{c|}{0.9719}        & \multicolumn{1}{c|}{0.9696}      & \multicolumn{1}{c|}{\textbf{0.9731}}        & 0.9773       \\ \cline{2-8} 
                      & GPU days                 & \multicolumn{1}{c|}{0.0068}       & \multicolumn{1}{c|}{7.4466}    & \multicolumn{1}{c|}{0.3412}       & \multicolumn{1}{c|}{0.0108}     & \multicolumn{1}{c|}{0.1084}       & -           \\ \hline
                      \hline
\multirow{4}{*}{REC.}   & Band Comb.          & \multicolumn{1}{c|}{8, 15, 20}    & \multicolumn{1}{c|}{7, 19, 24} & \multicolumn{1}{c|}{5, 16, 24}    & \multicolumn{1}{c|}{-}          & \multicolumn{1}{c|}{8, 15, 26}    & 7, 17, 27   \\ \cline{2-8} 
                      & MRAE                     & \multicolumn{1}{c|}{0.0605}       & \multicolumn{1}{c|}{0.0616}    & \multicolumn{1}{c|}{\textbf{0.0467}}       & \multicolumn{1}{c|}{-}          & \multicolumn{1}{c|}{0.0525}       & 0.0509      \\ \cline{2-8} 
                      & PSNR*                    & \multicolumn{1}{c|}{39.76}        & \multicolumn{1}{c|}{42.73}     & \multicolumn{1}{c|}{43.63}        & \multicolumn{1}{c|}{-}          & \multicolumn{1}{c|}{\textbf{43.79}}        & 44.32       \\ \cline{2-8} 
                      & GPU days                 & \multicolumn{1}{c|}{0.0534}       & \multicolumn{1}{c|}{24.60}     & \multicolumn{1}{c|}{2.6739}       & \multicolumn{1}{c|}{-}          & \multicolumn{1}{c|}{0.7008}       & -           \\ \hline
\end{tabular}
}
\caption{Main Results on our BSS-Bench. REC. denotes reconstruction and CLS. represents classification. The asterisk means the primary metric. Comb. denotes combination.}
\label{main_result}
\end{table*}

After that, 
the encoded MS input $M_H \in \mathbb{R}^{C\times HW}$ can be obtained by selecting $C = \{C_j, j\in [1, K]\}$ bands on $E_H$, where $K$ is the number of selected bands and $K$ satisfies $K < N$. 
%In our BSS-Bench, $K$ is set to $3$ by default.

We compare the effectiveness of these three encoding ways in the ablation study (see Table \ref{ablation_study}). The results show that the learnable position embedding is better and, with SLPE, our SCOS achieves the best performances.
The subsequent band selection search process is flexible and any adequate search algorithm \cite{Microsoft_Neural_Network_Intelligence_2021} is feasible. 
Search can be performed many times on the same backbone once trained. 
We randomly sample $M$ BCs for testing and select the BC with the best performance on the validation set as formulated in Eq. (\ref{problem_2}).

\section{Experiments}
In this section, we first present the main comparison results on our BSS-Bench including both current methods and our SCOS. 
Then, we carry out various aspects of analysis on BS including the impact of band numbers, different architectures, and different learning targets.
At last, we show the ablation study on different SSPE ways.

\textbf{Implementation details of BSS-Bench.} 
Important training hyper-parameters like metrics have been listed in Table \ref{bench_intro_table}. As for other details on two tasks, we strictly follow recent works.
For HSI reconstruction, we adopt the training and evaluation setting of MST++ \cite{cai2022mst++} such as the loss function and the batch size.
For HSI classification, we exploit the training and evaluation setting of SSFTT \cite{sun2022spectral} such as the loss function and the batch size.
The training of backbones does not exploit pre-trained weights by default.
All experiments are carried out on the same environment. 
%Our GPU server has two processors (Intel Xeon Gold 6238R) and ten graphic cards (GTX 3090).
The best BC and its performances on different tasks and backbones are also provided for reference and is named \textit{Oracle}. 
The Oracle on each task is the BC with the best performance on the primary metric. The primary metric of HSI classification is OA, and the primary metric of HSI reconstruction is PSNR.
For more details about our codes and BSS-Bench, we suggest that reviewers refer to our supplementary materials.

\textbf{The details of our SCOS.}
For the training on the HSI classification task, our SCOS adopts SSFTT \cite{sun2022spectral} as the backbone, which is the same as the transformer backbone used by our BSS-Bench (see Table \ref{bench_intro_table}) except for more training iterations.
The iterations of training SCOS on HSI classification is $320$k. 
Other training settings strictly follow SSFTT \cite{sun2022spectral}.
For the training on the HSI reconstruction task, our SCOS adopts MST++ \cite{cai2022mst++} as the backbone, which is the same as the transformer backbone adopted by our BSS-Bench (see Table \ref{bench_intro_table}). 
The iterations of training SCOS is 50k as we found through experiments that 50k is enough for converge on HSI reconstruction.
Other training settings are exactly the same as MST++ \cite{cai2022mst++}.
As for the search process, for HSI reconstruction, we randomly sample $M=1,000$ BCs to test their performance on the corresponding validation set. For HSI classification, the value of $M$ is set to $10,000$.
The BC with the top performance on the primary metric is regarded as the search result. After the BC is obtained, we finetune the backbone on it to obtain its exact performance following the setting in Table \ref{bench_intro_table}.
% search
% re-train 

\begin{table*}[!t]
\centering
\resizebox{0.99\linewidth}{!}{%
\begin{tabular}{c|ccccccccc|ccc}
\hline
Num of bands & \multicolumn{9}{c|}{full(200)}                                                      & 3         & 5         & 8        \\ \hline
No. of classes      & SVM    & EMAP   & 1D-CNN & 2D-CNN & 3D-CNN & SSRN   & Cubic-CNN & HybridSN & SSFTT  & \multicolumn{3}{c}{SSFTT + SCOS} \\ \hline
1            & 65.63  & 62.50  & 43.75  & 48.78  & 41.46  & 83.15  & 87.86     & 87.80    & 95.12  & 95.12     & 100.00    & 100.00   \\
2            & 63.44  & 81.57  & 77.93  & 78.13  & 90.51  & 95.31  & 96.35     & 94.39    & 97.67  & 96.18     & 94.78     & 95.48    \\
3            & 60.25  & 83.19  & 56.72  & 83.51  & 79.36  & 94.23  & 93.65     & 96.52    & 98.87  & 95.58     & 97.72     & 97.99    \\
4            & 41.11  & 85.89  & 45.18  & 47.42  & 46.01  & 90.68  & 82.54     & 83.89    & 91.55  & 95.77     & 94.83     & 96.71    \\
5            & 87.05  & 78.61  & 87.57  & 75.12  & 95.17  & 97.79  & 96.69     & 98.16    & 96.32  & 95.86     & 99.31     & 99.31    \\
6            & 97.21  & 79.08  & 98.63  & 92.99  & 99.70  & 98.67  & 96.69     & 99.54    & 99.54  & 99.54     & 99.23     & 99.54    \\
7            & 89.47  & 52.63  & 65.11  & 60.00  & 88.00  & 97.92  & 90.16     & 92.97    & 100.00 & 100.00    & 100.00    & 100.00   \\
8            & 96.66  & 91.19  & 97.36  & 98.37  & 100.00 & 99.26  & 98.46     & 100.00   & 100.00 & 100.00    & 100.00    & 100.00   \\
9            & 32.26  & 50.12  & 37.14  & 66.67  & 48.89  & 89.49  & 89.93     & 86.27    & 88.89  & 72.22     & 100.00    & 94.44    \\
10           & 73.84  & 81.32  & 66.03  & 87.77  & 86.06  & 97.48  & 93.94     & 97.94    & 97.71  & 97.71     & 94.40     & 98.28    \\
11           & 84.36  & 86.91  & 82.49  & 89.09  & 97.51  & 98.16  & 97.45     & 99.50    & 98.69  & 99.00     & 98.32     & 98.00    \\
12           & 42.89  & 78.43  & 73.49  & 63.67  & 74.91  & 93.07  & 93.18     & 94.57    & 98.13  & 95.31     & 97.19     & 97.19    \\
13           & 98.58  & 96.35  & 99.30  & 100.00 & 99.46  & 98.59  & 99.12     & 94.59    & 97.28  & 93.51     & 100.00    & 95.67    \\
14           & 94.02  & 93.91  & 93.78  & 95.33  & 99.74  & 99.72  & 99.39     & 99.29    & 99.91  & 100.00    & 99.82     & 99.47    \\
15           & 42.65  & 77.36  & 55.39  & 66.76  & 84.10  & 93.31  & 84.26     & 92.35    & 98.84  & 96.54     & 99.42     & 96.54    \\
16           & 92.19  & 84.38  & 81.54  & 91.57  & 93.98  & 93.79  & 89.69     & 97.98    & 95.54  & 91.66     & 89.28     & 86.90    \\ \hline
OA (\%)          & 76.39  & 83.69  & 79.37  & 84.47  & 91.03  & 94.78  & 94.90     & 96.62    & 97.47  & \textbf{97.64}     & 97.64     & 97.85    \\
AA (\%)          & 72.18  & 76.53  & 70.87  & 77.83  & 82.18  & 94.67  & 93.85     & 95.66    & 96.57  & 95.25     & \textbf{97.77}     & 97.22    \\
Kappa        & 0.7285 & 0.8140 & 0.7628 & 0.8224 & 0.8968 & 0.9408 & 0.9417    & 0.9629   & 0.9711 & \textbf{0.9731}    & 0.9731    & 0.9755   \\ \hline
\end{tabular}
}
\caption{Comparisons between SOTA methods and our SCOS on the Indian Pines dataset. 
Our SCOS adopts the same backbone as SSFTT. We bold the values where SCOS surpasses current full-band methods with the minimum number of bands.
}
\label{ip_compare}
\end{table*}

\textbf{The main result on our BSS-Bench.} We compare recent works on BSS-Bench: IBRA \cite{morales2021hyperspectral}, Predictor \cite{wen2020neural}, MHRO \cite{ye2022band}, DRL \cite{mou2021deep}. The implementation details of Predictor is available in the supplementary material.
Metrics and the cost measured by GPU days are list in Table \ref{main_result}.
In comparison to unsupervised BS techniques such as IBRA, the genetic algorithm MHRO generates superior band combinations by maintaining diverse populations and utilizing evolution operators for iterative searching. 
Though better band combinations are found, the computational cost of $0.3412$ GPU days is considerable.
The excellent AA performance of DRL implies that reinforcement learning could be a promising approach for BS by framing it as a Markov decision process.
Learning-based methods like Predictor randomly samples a subset of the entire search space, train a model, and exploit it to predict the performance of any BCs. The following search process is similar to our SCOS. 
Though achieving better performances, Predictor is computationally intensive. It needs $7.4466$ GPU days for HSI classification and $24.60$ GPU days for HSI reconstruction.
In contrast, our SCOS only needs one-time training, reducing the total GPU days significantly to about one-fiftieth of Predictor.
Furthermore, our SCOS achieves the best performances on all metrics. 
Compared to the Oracle result that represents the best band combination in our BSS-Bench, our SCOS reduces the gap between current methods and the optimal result in terms of Kappa by $22$\%, demonstrating its effectiveness.

%we find that the unsupervised method DRL achieves the best performance on AA.
%DRL frame the problem of unsupervised band selection as a Markov decision process, propose an effective method to parameterize it, and finally solve the problem by deep reinforcement learning.

\textbf{The impact of the number of bands.} Is the performance of spectral analysis positively correlated with the number of bands? 
To answer this question, we compare recent works on HSI classification using full bands in Table \ref{ip_compare}: SVM, EMAP \cite{dalla2010classification}, 1D-CNN \cite{hu2015deep}, 2D-CNN, \cite{zhao2016spectral}, 3D-CNN \cite{chen2016deep}, SSRN \cite{zhong2017spectral}, Cubic-CNN \cite{wang2020novel}, HybridSN \cite{roy2019hybridsn}, SSFTT \cite{sun2022spectral}. 
Our findings indicate that the number of bands does not always directly correlate with performance.
%We find that the performance is not necessarily proportional to the number of bands. 
We are able to achieve competitive results on OA ($97.64$\%) and Kappa ($0.9731$) with our SCOS, using only three carefully selected bands $(92, 121, 162)$. This outperforms the current state-of-the-art methods that use all $200$ bands on both OA and Kappa. 
As for AA, our SCOS outperforms current methods using only one-fortieth ($5$ bands) the number of bands of current state-of-the-art methods.
%With only three selected bands (92, 121, 162), our SCOS achieves very competitive results on OA (97.64\%) and Kappa (0.9731), surpassing state-of-the-art methods with the full 200 bands. 
%Note that our SCOS exploits the same backbone as SSFTT. By selecting effective band combinations, SCOS achieves the best performance with only three bands than SSFTT with 200 bands. 
It is important to note that SCOS and SSFTT share the same backbone. By selecting effective band combinations, SCOS achieves higher performances on OA and Kappa with just three bands.
This remarkable performance contrast highlights the high redundancy of features in HSI data and emphasizes the importance of effective BS algorithms.
As shown in Table \ref{main_result}, not all BCs achieve state-of-the-art performances. Fig. \ref{intro_scatter} also shows that the performances of different band combinations vary significantly.
Therefore, the superior performance on three bands also suggests the effectiveness of our SCOS.

\begin{figure}[!t]
\centering
\includegraphics[width=0.49\linewidth]{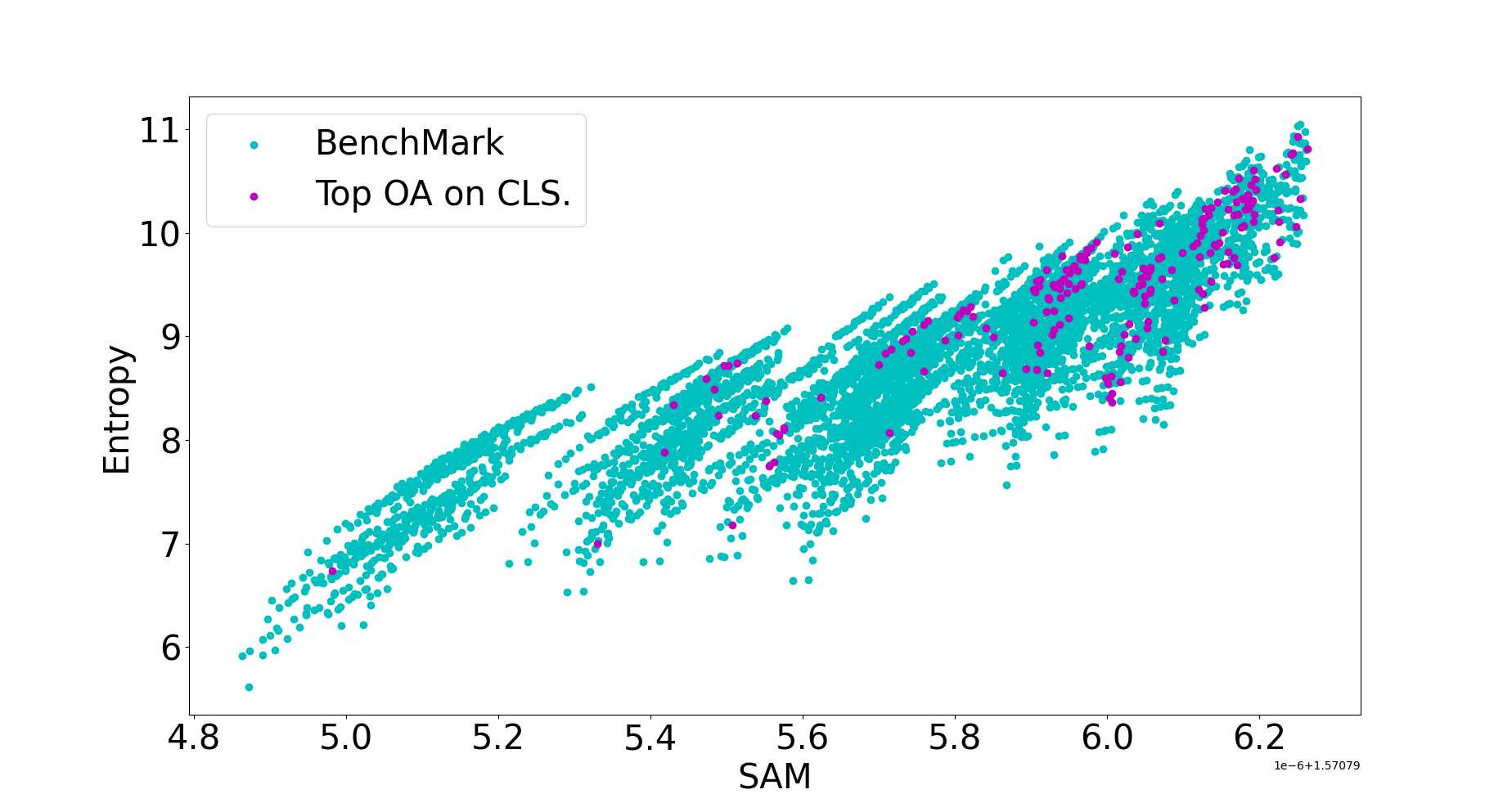}
\includegraphics[width=0.49\linewidth]{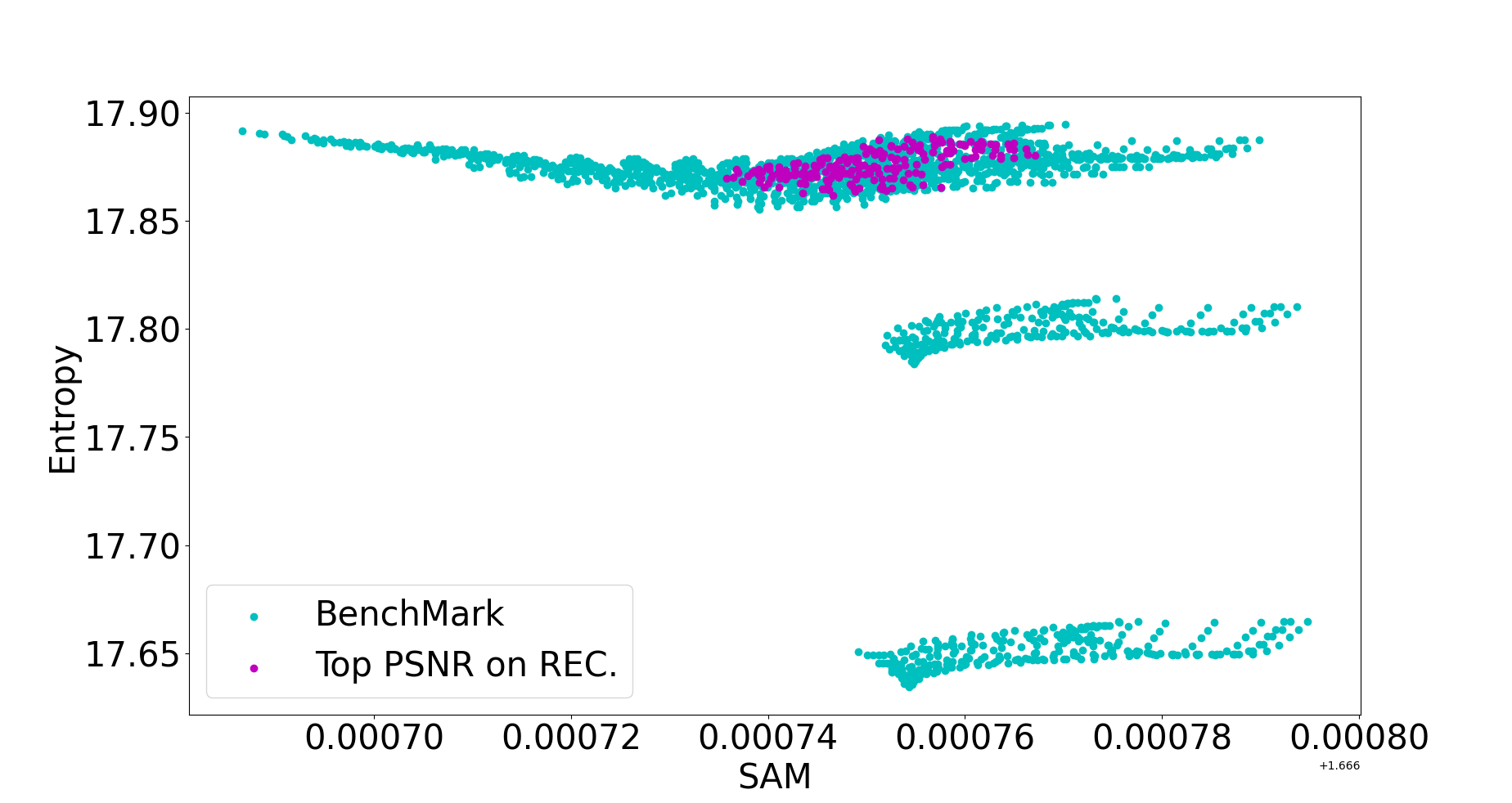}
\caption{The effectiveness of two unsupervised statistics (Entropy and SAM) on HSI classification (left) and HSI reconstruction (right).
Best viewed in color.
}
\label{unsupervised_statistic_compare}
\end{figure}

\textbf{The effectiveness of unsupervised statistics.} Can unsupervised methods find good BCs? 
To answer this question, we plot the distributions of all BCs based on two well-established unsupervised metrics SAM \cite{chakravarty2021hyperspectral} and Entropy \cite{morales2021hyperspectral,mou2021deep}. The entropy of a BC is obtained by averaging the entropy of each band within that BC. The SAM of a BC is computed by averaging the SAM of all possible band pairs within that BC.
As shown in Fig. \ref{unsupervised_statistic_compare}, each point represents a BC from our BSS-Bench and we mark the top $5$\% BCs in purple. The distribution of purple points shows that unsupervised statistics can find good BCs as most top BCs have higher entropy value and higher SAM value. At the same time, not all BCs with higher entropy values and higher SAM values achieve high performances. Therefore, unsupervised statistics may serve as effective initialization methods for BS but might struggle to identify the best BC.
The subsequent experiment further confirms that it is difficult for unsupervised methods to find the best BCs.

\begin{figure}[!t]
\centering
\includegraphics[width=0.49\linewidth]{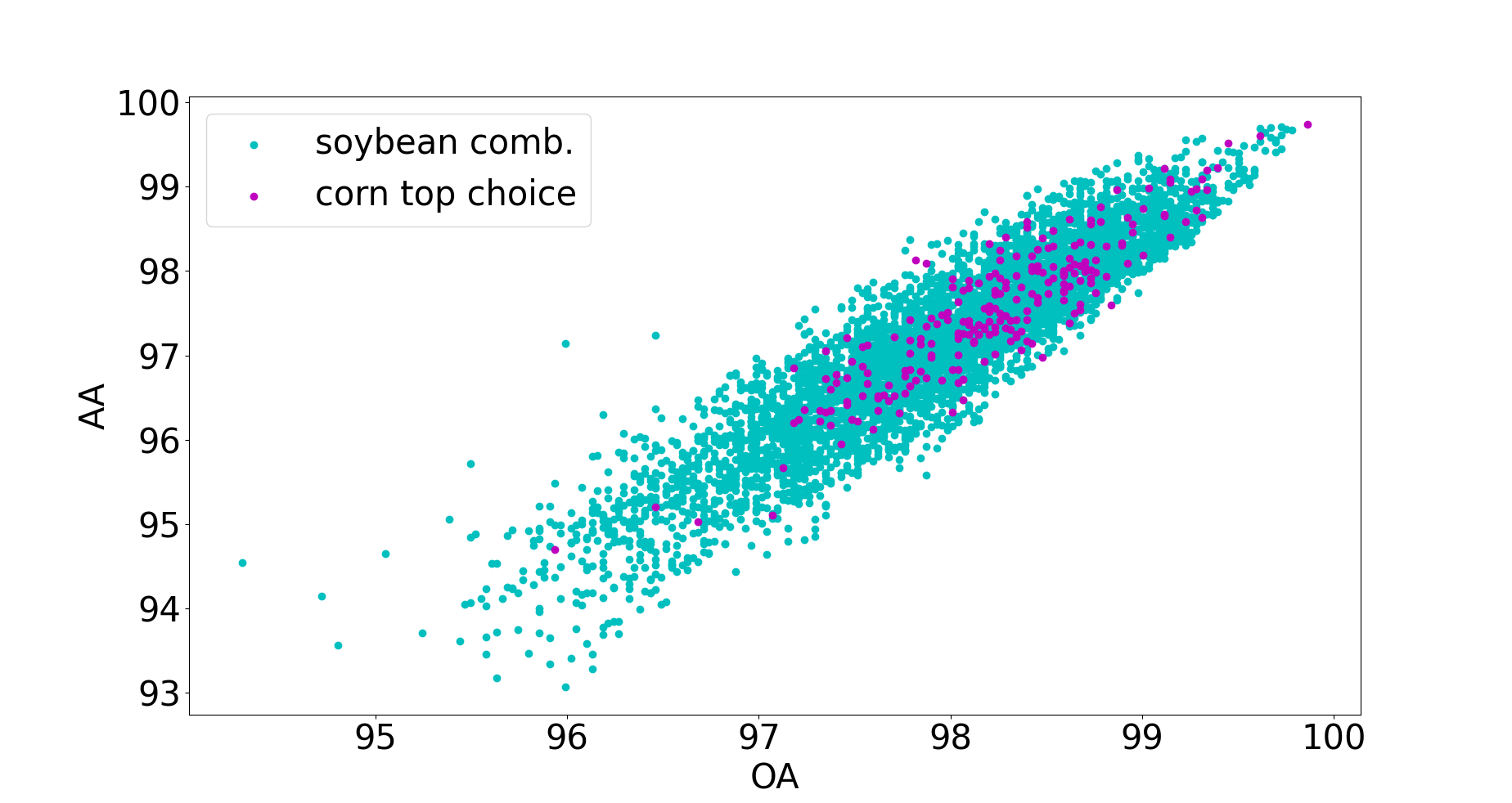}
\includegraphics[width=0.49\linewidth]{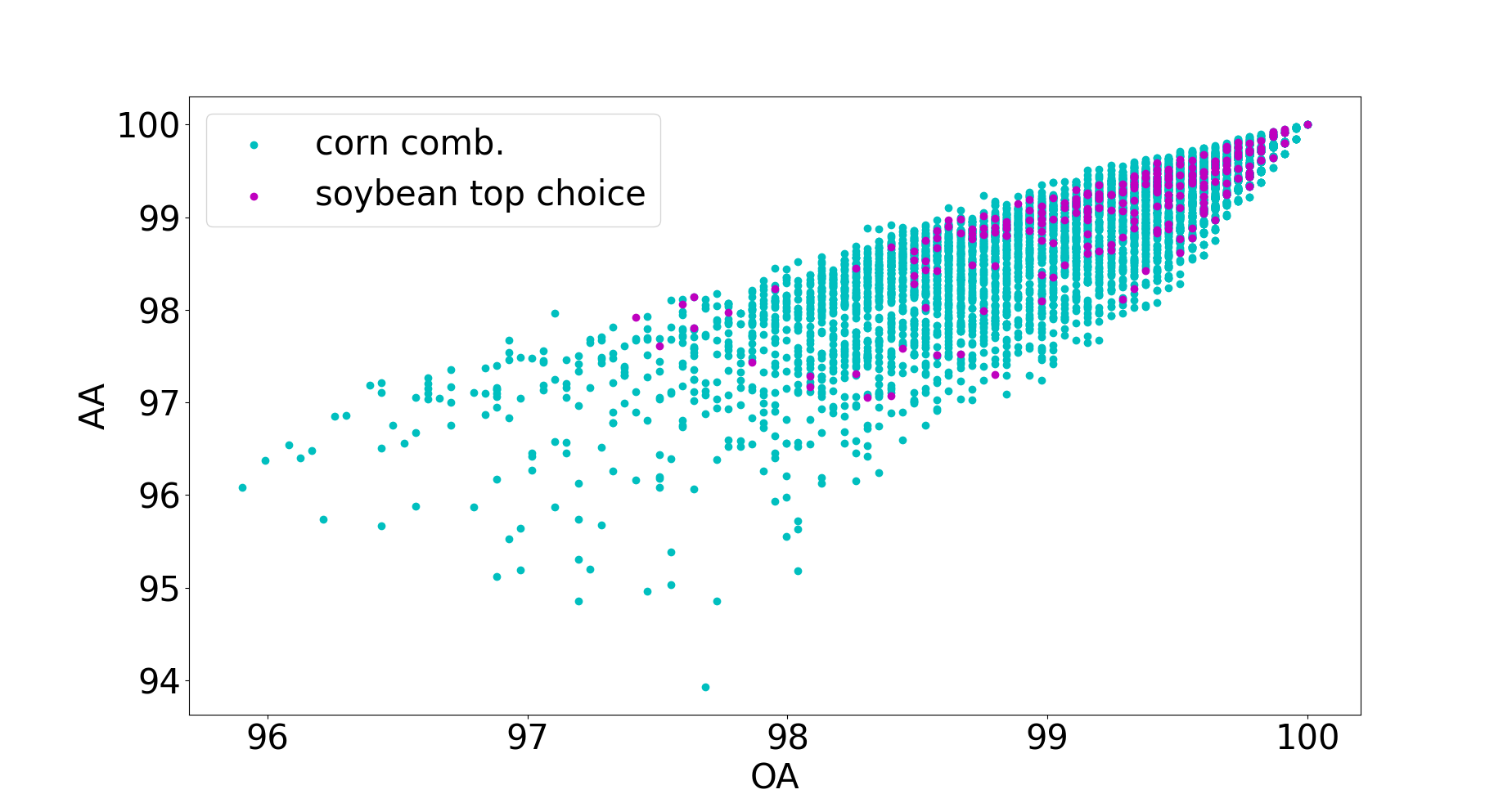}
\caption{The impact of two different HSI optimization targets on top band combinations. The distribution of top band combinations for different optimization targets varies significantly.
Best viewed in color.
}
\label{different_target_impact}
\end{figure}

\textbf{The impact of different optimization targets.}
Will top BCs change with different optimization targets?
To answer this question, we split labels of the Indian Pines dataset into two disjoint sub-tasks: HSI corn classifications ($3$ classes with corn in their names) and HSI soybean classifications ($3$ classes with soybean in their names). On each sub-task, other classes are all considered as the background. Then, we construct the performance table for each sub-task according to the Step 3 and the Step 4 described in Section \ref{bench_section} and obtain the top BCs for each sub-task. The results are plotted in Fig. \ref{different_target_impact}.
On the left subfigure, we plot the distribution of all BCs for soybean classification and mark the top BCs of the corn classification sub-task in purple. Similarly, on the right subfigure, we plot the distribution of all BCs for the corn classification sub-task and mark the top BCs of the soybean classification sub-task in purple. 
The results revealed that the top BCs for the corn classification sub-task do not necessarily perform well in the soybean classification sub-task, and vice versa.
As the dataset is exactly the same, unsupervised methods will select the same BC for different HSI analysis targets. However, different materials have different physical footprints, resulting in potentially different top-performance BCs. 
Therefore, learning-based methods may select better BCs than unsupervised strategies based on different optimization targets.

\begin{figure}[!t]
\centering
\includegraphics[width=0.49\linewidth]{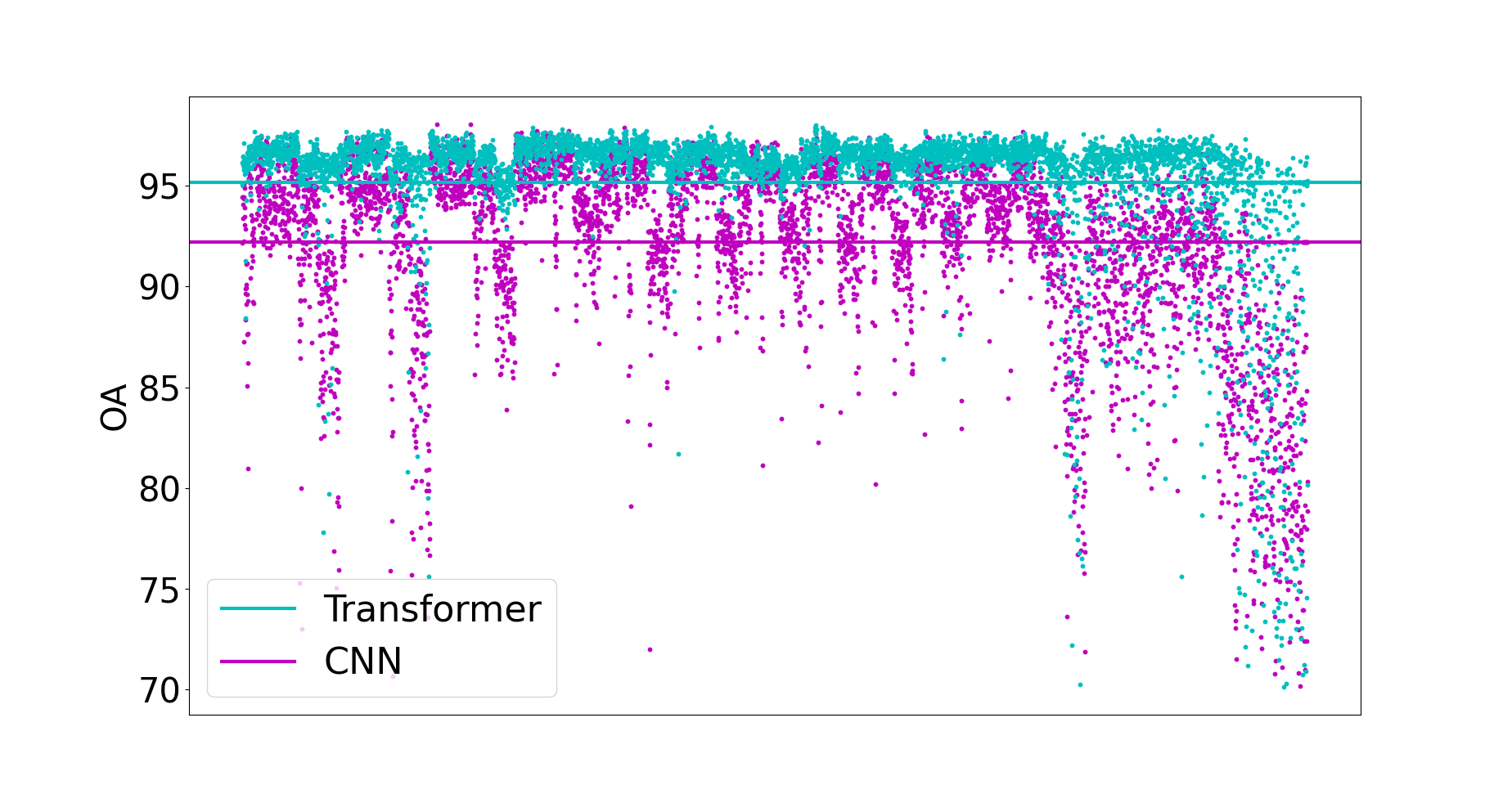}
\includegraphics[width=0.49\linewidth]{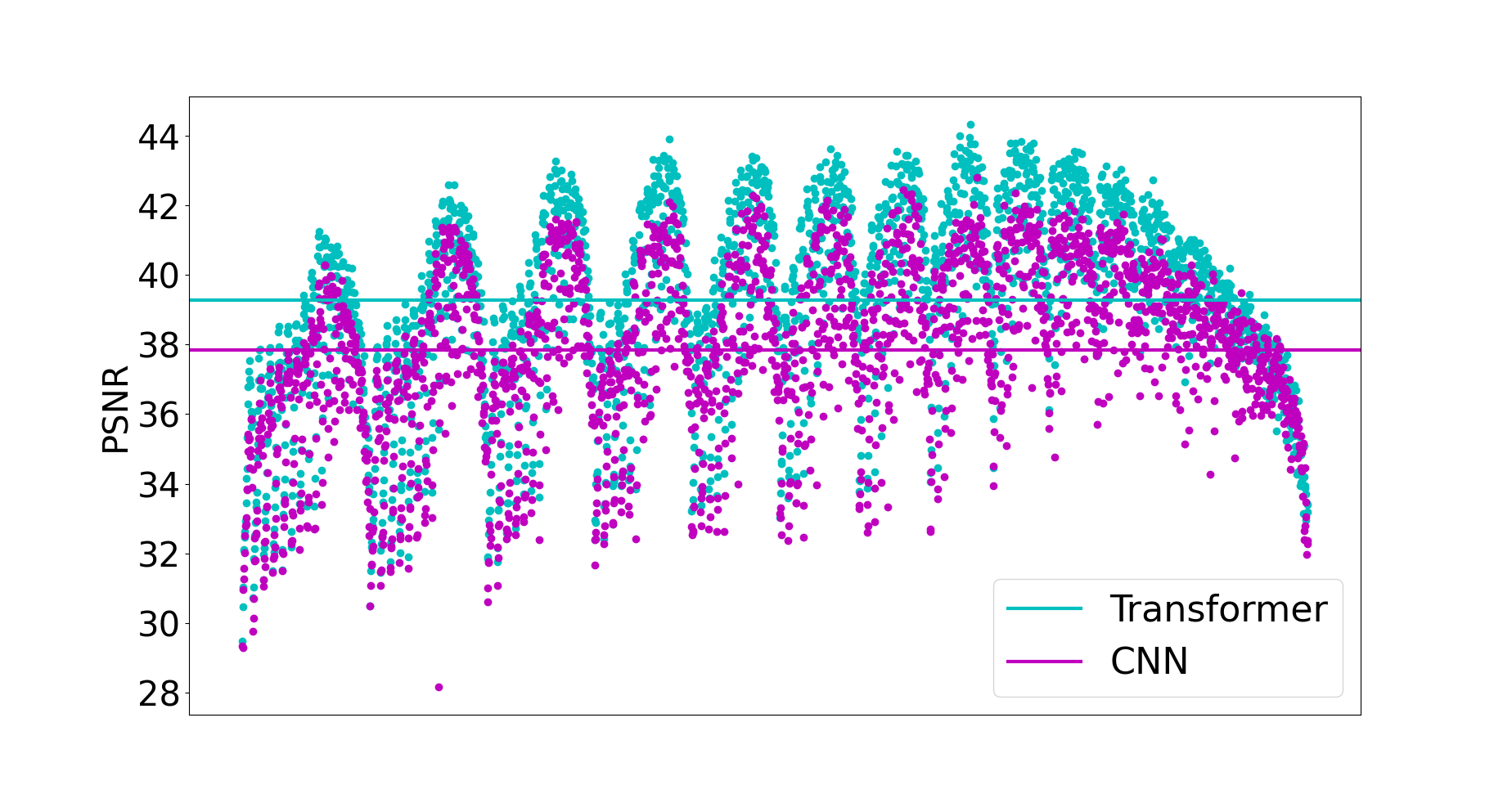}
\caption{The impact of two different architectures on BS. We sort all band combinations by band indexes and plot the performances of Transformer (in turquoise blue) and CNN (in purple) on the HSI classification task (left) and the HSI reconstruction task (right).
Best viewed in color.
}
\label{different_architecture_impact}
\end{figure}

\textbf{The effectiveness of different architectures.}
Will the superiority of the BC be affected by different backbones?
To answer this question, we arranged all BCs by their band indexes and plot the performances of both CNN and Transformer in Fig. \ref{different_architecture_impact}. Results show that the performance changes consistently with changing band indexes.
Additionally, it is noticeable that, in Fig. \ref{different_architecture_impact}, the top BCs based on the CNN architecture are heavily overlapped with the top BCs based on the transformer architecture.
This ranking consistency confirms that the effectiveness of BCs is not significantly impacted by variations in backbones.
%as for the HSI reconstruction task in our BSS-Bench,  the best BC found by the CNN architecture ranks $2$-th in the table of the Transformer architecture. 
% may be caused by the randomness of the training, which is a normal phenomenon.

\begin{table}[!t]
\centering
\resizebox{0.9\linewidth}{!}{%
\begin{tabular}{|l|l|l|l||l|}
\hline
Method      & APE     & CLPE    & SLPE    & Oracle  \\ \hline
MRAE      & 0.063   & 0.056   & \textbf{0.053}   & 0.051   \\ \hline
PSNR      & 42.89   & 43.42   & \textbf{43.79}   & 44.32   \\ \hline
Band Comb. & 8,17,28 & 7,18,26 & \textbf{8,15,26} & 7,17,27 \\ \hline
\end{tabular}
}
\caption{Ablation study on different spectral spatial position embedding methods. APE denotes absolute position embedding. CLPE represents co-learnable position embedding and SLPE means separately learnable position embedding.
}
\label{ablation_study}
\end{table}

\textbf{Ablation study on SSPE.}
To compare different ways of spectral spatial position embeddings, we integrate three embedding methods introduced in Section \ref{method_section}, one at a time, with our SCOS and evaluate their search result on the HSI reconstruction task of BSS-Bench.
%in turn with our SCOS and compare their search results on the HSI reconstruction task of our BSS-Bench. 
The results are shown in Table \ref{ablation_study}.
The learnable position embedding exhibits better performance than the absolute position embedding. Furthermore, when position embedding is separately learnable, our SCOS achieves noticeable performance gains. 
Moreover, the discrepancy between the Oracle result and our SCOS is small ($0.53$ on PSNR and $0.002$ on MRAE), demonstrating the effectiveness of our proposed SSPE.

\section{Conclusions}
In this study, we introduced BSS-Bench, the first benchmark for band selection search, to facilitate reproducible research in the field of BS.
Using two types of backbones, we tested numerous band combinations on two distinct HSI analysis tasks, resulting in over $52$k records.
BSS-Bench enables BS methods to be performed by querying the benchmark table rather than carrying out the actual training and evaluation steps.
Furthermore, with the aid of BSS-Bench, we conducted a comprehensive analysis of the impact of different factors on BS. Three insights can be inferred:
\begin{itemize}[noitemsep,nolistsep]
    \item With effective BS algorithms, three bands are enough to achieve high HSI classification performances that is comparable to the state-of-the-art methods using hundreds of bands.
    \item Supervised learning-based BS methods have the potential to select better band combinations compared to unsupervised approaches.
    \item The superiority of different band combinations is less affected by different backbones.
\end{itemize}
Furthermore, to enable effective band selection, we introduce a novel method named Single Combination One Shot (SCOS), which learns the priority of band combinations through one-time training. The search step is training-free and flexible, allowing SCOS to be combined with various search algorithms. Evaluations on different spectral analysis tasks demonstrate that our SCOS outperforms current HSI classification methods using only three bands.

\textbf{Limitations.}
The construction of BSS-Bench has improved the replicability of BS research and alleviated the computational burden of large-scale experiments. However, we acknowledge three limitations to our benchmark.
Firstly, some popular BS datasets such as the Pavia University and the Houston $2013$ datasets are not supported currently.
Due to the significant amount of computational resources required, adding a new dataset is a time-consuming process. 
We are continually enriching the number of data sets and the number of tasks.
Secondly, BSS-Bench does not answer the relation between the BS performance and the number of bands as the default number of bands is fixed to three. Our evaluations have shown that this correlation is not necessarily positive, and we plan to investigate their relationship in future work.
Thirdly, while we have invested a significant amount of effort in implementing existing methods on the setting of BSS-Bench, some algorithms may not produce optimal results as their hyper-parameters may not align with our default settings. We will update the benchmark results if researchers provide compelling evidence of improved performance using different hyper-parameters.
%As adding a dataset may need thousands of GPU days, it takes a certain amount of time.
%In this paper, we proposed the first band selection search benchmark named BSS-Bench to make BS researches easy and reproducible. 
%Numerous band combinations on two kinds of backbones and two different HSI analysis tasks are tested, producing over 52k records.
%We have tried our best to implement each method. However, still, some algorithms might obtain non-optimal results since their hyper-parameters might not fit our

{\small
\bibliographystyle{ieee_fullname}
\bibliography{egbib}

\begin{thebibliography}{10}\itemsep=-1pt

\bibitem{al2020review}
Qasem Al-Tashi, Helmi Md~Rais, Said~Jadid Abdulkadir, Seyedali Mirjalili, and Hitham Alhussian.
\newblock A review of grey wolf optimizer-based feature selection methods for classification.
\newblock {\em Evolutionary Machine Learning Techniques: Algorithms and Applications}, pages 273--286, 2020.

\bibitem{alkhatib2023tri}
Mohammed~Q Alkhatib, Mina Al-Saad, Nour Aburaed, Saeed Almansoori, Jaime Zabalza, Stephen Marshall, and Hussain Al-Ahmad.
\newblock Tri-cnn: a three branch model for hyperspectral image classification.
\newblock {\em Remote Sensing}, 15(2):316, 2023.

\bibitem{arad2022ntire}
Boaz Arad, Radu Timofte, Rony Yahel, Nimrod Morag, Amir Bernat, Yuanhao Cai, Jing Lin, Zudi Lin, Haoqian Wang, Yulun Zhang, et~al.
\newblock Ntire 2022 spectral recovery challenge and data set.
\newblock In {\em Proceedings of the IEEE/CVF Conference on Computer Vision and Pattern Recognition}, pages 863--881, 2022.

\bibitem{baumgardner2015220}
Marion~F Baumgardner, Larry~L Biehl, and David~A Landgrebe.
\newblock 220 band aviris hyperspectral image data set: June 12, 1992 indian pine test site 3.
\newblock {\em Purdue University Research Repository}, 10(7):991, 2015.

\bibitem{cai2022mst++}
Yuanhao Cai, Jing Lin, Zudi Lin, Haoqian Wang, Yulun Zhang, Hanspeter Pfister, Radu Timofte, and Luc Van~Gool.
\newblock Mst++: Multi-stage spectral-wise transformer for efficient spectral reconstruction.
\newblock In {\em Proceedings of the IEEE/CVF Conference on Computer Vision and Pattern Recognition}, pages 745--755, 2022.

\bibitem{dauhst}
Yuanhao Cai, Jing Lin, Haoqian Wang, Xin Yuan, Henghui Ding, Yulun Zhang, Radu Timofte, and Luc Van~Gool.
\newblock Degradation-aware unfolding half-shuffle transformer for spectral compressive imaging.
\newblock In {\em NeurIPS}, 2022.

\bibitem{chakravarty2021hyperspectral}
Sujata Chakravarty, Bijay~Kumar Paikaray, Rutuparnna Mishra, and Satyabrata Dash.
\newblock Hyperspectral image classification using spectral angle mapper.
\newblock In {\em 2021 IEEE International Women in Engineering (WIE) Conference on Electrical and Computer Engineering (WIECON-ECE)}, pages 87--90. IEEE, 2021.

\bibitem{chen2021hinet}
Liangyu Chen, Xin Lu, Jie Zhang, Xiaojie Chu, and Chengpeng Chen.
\newblock Hinet: Half instance normalization network for image restoration.
\newblock In {\em Proceedings of the IEEE/CVF Conference on Computer Vision and Pattern Recognition}, pages 182--192, 2021.

\bibitem{chen2016deep}
Yushi Chen, Hanlu Jiang, Chunyang Li, Xiuping Jia, and Pedram Ghamisi.
\newblock Deep feature extraction and classification of hyperspectral images based on convolutional neural networks.
\newblock {\em IEEE Transactions on Geoscience and Remote Sensing}, 54(10):6232--6251, 2016.

\bibitem{dalla2010classification}
Mauro Dalla~Mura, Alberto Villa, Jon~Atli Benediktsson, Jocelyn Chanussot, and Lorenzo Bruzzone.
\newblock Classification of hyperspectral images by using extended morphological attribute profiles and independent component analysis.
\newblock {\em IEEE Geoscience and Remote Sensing Letters}, 8(3):542--546, 2010.

\bibitem{ding2020improved}
Xiaohui Ding, Huapeng Li, Ji Yang, Patricia Dale, Xiangcong Chen, Chunlei Jiang, and Shuqing Zhang.
\newblock An improved ant colony algorithm for optimized band selection of hyperspectral remotely sensed imagery.
\newblock {\em IEEE Access}, 8:25789--25799, 2020.

\bibitem{dong2020bench}
Xuanyi Dong and Yi Yang.
\newblock Nas-bench-201: Extending the scope of reproducible neural architecture search.
\newblock {\em arXiv preprint arXiv:2001.00326}, 2020.

\bibitem{dosovitskiy2020image}
Alexey Dosovitskiy, Lucas Beyer, Alexander Kolesnikov, Dirk Weissenborn, Xiaohua Zhai, Thomas Unterthiner, Mostafa Dehghani, Matthias Minderer, Georg Heigold, Sylvain Gelly, et~al.
\newblock An image is worth 16x16 words: Transformers for image recognition at scale.
\newblock {\em arXiv preprint arXiv:2010.11929}, 2020.

\bibitem{driess2023palme}
Danny Driess, Fei Xia, Mehdi S.~M. Sajjadi, Corey Lynch, Aakanksha Chowdhery, Brian Ichter, Ayzaan Wahid, Jonathan Tompson, Quan Vuong, Tianhe Yu, Wenlong Huang, Yevgen Chebotar, Pierre Sermanet, Daniel Duckworth, Sergey Levine, Vincent Vanhoucke, Karol Hausman, Marc Toussaint, Klaus Greff, Andy Zeng, Igor Mordatch, and Pete Florence.
\newblock Palm-e: An embodied multimodal language model.
\newblock In {\em arXiv preprint arXiv:2303.03378}, 2023.

\bibitem{feng2021deep}
Jie Feng, Di Li, Jing Gu, Xianghai Cao, Ronghua Shang, Xiangrong Zhang, and Licheng Jiao.
\newblock Deep reinforcement learning for semisupervised hyperspectral band selection.
\newblock {\em IEEE Transactions on Geoscience and Remote Sensing}, 60:1--19, 2021.

\bibitem{grewal2023machine}
Reaya Grewal, Singara Singh~Kasana, and Geeta Kasana.
\newblock Machine learning and deep learning techniques for spectral spatial classification of hyperspectral images: A comprehensive survey.
\newblock {\em Electronics}, 12(3):488, 2023.

\bibitem{gu2017multiple}
Yanfeng Gu, Jocelyn Chanussot, Xiuping Jia, and Jon~Atli Benediktsson.
\newblock Multiple kernel learning for hyperspectral image classification: A review.
\newblock {\em IEEE Transactions on Geoscience and Remote Sensing}, 55(11):6547--6565, 2017.

\bibitem{hu2015deep}
Wei Hu, Yangyu Huang, Li Wei, Fan Zhang, and Hengchao Li.
\newblock Deep convolutional neural networks for hyperspectral image classification.
\newblock {\em Journal of Sensors}, 2015:1--12, 2015.

\bibitem{hu2022hdnet}
Xiaowan Hu, Yuanhao Cai, Jing Lin, Haoqian Wang, Xin Yuan, Yulun Zhang, Radu Timofte, and Luc Van~Gool.
\newblock Hdnet: High-resolution dual-domain learning for spectral compressive imaging.
\newblock In {\em Proceedings of the IEEE/CVF Conference on Computer Vision and Pattern Recognition}, pages 17542--17551, 2022.

\bibitem{huang2021deep}
Tao Huang, Weisheng Dong, Xin Yuan, Jinjian Wu, and Guangming Shi.
\newblock Deep gaussian scale mixture prior for spectral compressive imaging.
\newblock In {\em Proceedings of the IEEE/CVF Conference on Computer Vision and Pattern Recognition}, pages 16216--16225, 2021.

\bibitem{li2021improved}
An-Da Li, Bing Xue, and Mengjie Zhang.
\newblock Improved binary particle swarm optimization for feature selection with new initialization and search space reduction strategies.
\newblock {\em Applied Soft Computing}, 106:107302, 2021.

\bibitem{li2023jointly}
Ke Li, Dengxin Dai, and Luc Van~Gool.
\newblock Jointly learning band selection and filter array design for hyperspectral imaging.
\newblock In {\em Proceedings of the IEEE/CVF Winter Conference on Applications of Computer Vision}, pages 6384--6394, 2023.

\bibitem{li2023hyperspectral}
Shuying Li, Baidong Peng, Long Fang, Qiang Zhang, Lei Cheng, and Qiang Li.
\newblock Hyperspectral band selection via difference between inter-groups.
\newblock {\em IEEE Transactions on Geoscience and Remote Sensing}, 2023.

\bibitem{liu2018rank}
Yang Liu, Xin Yuan, Jinli Suo, David~J Brady, and Qionghai Dai.
\newblock Rank minimization for snapshot compressive imaging.
\newblock {\em IEEE transactions on pattern analysis and machine intelligence}, 41(12):2990--3006, 2018.

\bibitem{meng2021self}
Ziyi Meng, Zhenming Yu, Kun Xu, and Xin Yuan.
\newblock Self-supervised neural networks for spectral snapshot compressive imaging.
\newblock In {\em Proceedings of the IEEE/CVF International Conference on Computer Vision}, pages 2622--2631, 2021.

\bibitem{miao2019net}
Xin Miao, Xin Yuan, Yunchen Pu, and Vassilis Athitsos.
\newblock l-net: Reconstruct hyperspectral images from a snapshot measurement.
\newblock In {\em Proceedings of the IEEE/CVF International Conference on Computer Vision}, pages 4059--4069, 2019.

\bibitem{Microsoft_Neural_Network_Intelligence_2021}
{Microsoft}.
\newblock {Neural Network Intelligence}, 1 2021.

\bibitem{mills2022aio}
Keith~G Mills, Di Niu, Mohammad Salameh, Weichen Qiu, Fred~X Han, Puyuan Liu, Jialin Zhang, Wei Lu, and Shangling Jui.
\newblock Aio-p: Expanding neural performance predictors beyond image classification.
\newblock {\em arXiv preprint arXiv:2211.17228}, 2022.

\bibitem{morales2021hyperspectral}
Giorgio Morales, John Sheppard, Riley Logan, and Joseph Shaw.
\newblock Hyperspectral band selection for multispectral image classification with convolutional networks.
\newblock In {\em 2021 International Joint Conference on Neural Networks (IJCNN)}, pages 1--8. IEEE, 2021.

\bibitem{mou2021deep}
Lichao Mou, Sudipan Saha, Yuansheng Hua, Francesca Bovolo, Lorenzo Bruzzone, and Xiao~Xiang Zhu.
\newblock Deep reinforcement learning for band selection in hyperspectral image classification.
\newblock {\em IEEE Transactions on Geoscience and Remote Sensing}, 60:1--14, 2021.

\bibitem{qiao2020deep}
Mu Qiao, Ziyi Meng, Jiawei Ma, and Xin Yuan.
\newblock Deep learning for video compressive sensing.
\newblock {\em Apl Photonics}, 5(3):030801, 2020.

\bibitem{ramesh2022hierarchical}
Aditya Ramesh, Prafulla Dhariwal, Alex Nichol, Casey Chu, and Mark Chen.
\newblock Hierarchical text-conditional image generation with clip latents.
\newblock {\em arXiv preprint arXiv:2204.06125}, 2022.

\bibitem{rani2022machine}
Alka Rani, Nirmal Kumar, Jitendra Kumar, and Nishant~K Sinha.
\newblock Machine learning for soil moisture assessment.
\newblock In {\em Deep Learning for Sustainable Agriculture}, pages 143--168. Elsevier, 2022.

\bibitem{rezasoltani2023hyperspectral}
Shima Rezasoltani and Faisal~Z Qureshi.
\newblock Hyperspectral image compression using implicit neural representation.
\newblock {\em arXiv preprint arXiv:2302.04129}, 2023.

\bibitem{roy2019hybridsn}
Swalpa~Kumar Roy, Gopal Krishna, Shiv~Ram Dubey, and Bidyut~B Chaudhuri.
\newblock Hybridsn: Exploring 3-d--2-d cnn feature hierarchy for hyperspectral image classification.
\newblock {\em IEEE Geoscience and Remote Sensing Letters}, 17(2):277--281, 2019.

\bibitem{sarhrouni2012band}
Elkebir Sarhrouni, Ahmed Hammouch, and Driss Aboutajdine.
\newblock Band selection and classification of hyperspectral images by minimizing normalized mutual information.
\newblock In {\em Second International Conference on the Innovative Computing Technology (INTECH 2012)}, pages 184--189. IEEE, 2012.

\bibitem{stuart2022high}
Mary~B Stuart, Matthew Davies, Matthew~J Hobbs, Tom~D Pering, Andrew~JS McGonigle, and Jon~R Willmott.
\newblock High-resolution hyperspectral imaging using low-cost components: Application within environmental monitoring scenarios.
\newblock {\em Sensors}, 22(12):4652, 2022.

\bibitem{su2018saliency}
Peifeng Su, Daizhi Liu, Xihai Li, and Zhigang Liu.
\newblock A saliency-based band selection approach for hyperspectral imagery inspired by scale selection.
\newblock {\em IEEE Geoscience and Remote Sensing Letters}, 15(4):572--576, 2018.

\bibitem{sun2022spectral}
Le Sun, Guangrui Zhao, Yuhui Zheng, and Zebin Wu.
\newblock Spectral--spatial feature tokenization transformer for hyperspectral image classification.
\newblock {\em IEEE Transactions on Geoscience and Remote Sensing}, 60:1--14, 2022.

\bibitem{sun2021multiscale}
Weiwei Sun, Gang Yang, Jiangtao Peng, Xiangchao Meng, Ke He, Wei Li, Heng-Chao Li, and Qian Du.
\newblock A multiscale spectral features graph fusion method for hyperspectral band selection.
\newblock {\em IEEE Transactions on Geoscience and Remote Sensing}, 60:1--12, 2021.

\bibitem{tang2021hyperspectral}
Chang Tang, Xinwang Liu, En Zhu, Lizhe Wang, and Albert~Y Zomaya.
\newblock Hyperspectral band selection via spatial-spectral weighted region-wise multiple graph fusion-based spectral clustering.
\newblock In {\em IJCAI}, pages 3038--3044, 2021.

\bibitem{wang2020novel}
Jinwei Wang, Xiangbo Song, Le Sun, Wei Huang, and Jin Wang.
\newblock A novel cubic convolutional neural network for hyperspectral image classification.
\newblock {\em IEEE Journal of Selected Topics in Applied Earth Observations and Remote Sensing}, 13:4133--4148, 2020.

\bibitem{wang2023simple}
Junyu Wang, Shijie Wang, Wenyu Liu, Zengqiang Zheng, and Xinggang Wang.
\newblock A simple adaptive unfolding network for hyperspectral image reconstruction.
\newblock {\em arXiv preprint arXiv:2301.10208}, 2023.

\bibitem{wen2020neural}
Wei Wen, Hanxiao Liu, Yiran Chen, Hai Li, Gabriel Bender, and Pieter-Jan Kindermans.
\newblock Neural predictor for neural architecture search.
\newblock In {\em Computer Vision--ECCV 2020: 16th European Conference, Glasgow, UK, August 23--28, 2020, Proceedings, Part XXIX}, pages 660--676. Springer, 2020.

\bibitem{xiao2022shapley}
Han Xiao, Ziwei Wang, Zheng Zhu, Jie Zhou, and Jiwen Lu.
\newblock Shapley-nas: Discovering operation contribution for neural architecture search.
\newblock In {\em Proceedings of the IEEE/CVF Conference on Computer Vision and Pattern Recognition}, pages 11892--11901, 2022.

\bibitem{xu2021similarity}
Buyun Xu, Xihai Li, Weijun Hou, Yiting Wang, and Yiwei Wei.
\newblock A similarity-based ranking method for hyperspectral band selection.
\newblock {\em IEEE Transactions on Geoscience and Remote Sensing}, 59(11):9585--9599, 2021.

\bibitem{yang2023double}
Hua Yang, Ming Chen, Guowen Wu, Jiali Wang, Yingxi Wang, and Zhonghua Hong.
\newblock Double deep q-network for hyperspectral image band selection in land cover classification applications.
\newblock {\em Remote Sensing}, 15(3):682, 2023.

\bibitem{ye2023beta}
Peng Ye, Tong He, Baopu Li, Tao Chen, Lei Bai, and Wanli Ouyang.
\newblock $\beta$-darts++: Bi-level regularization for proxy-robust differentiable architecture search.
\newblock {\em arXiv preprint arXiv:2301.06393}, 2023.

\bibitem{ye2022band}
Zhiwei Ye, Wenhui Cai, Shiqin Liu, Kainan Liu, Mingwei Wang, and Wen Zhou.
\newblock A band selection approach for hyperspectral image based on a modified hybrid rice optimization algorithm.
\newblock {\em Symmetry}, 14(7):1293, 2022.

\bibitem{ying2019bench}
Chris Ying, Aaron Klein, Eric Christiansen, Esteban Real, Kevin Murphy, and Frank Hutter.
\newblock Nas-bench-101: Towards reproducible neural architecture search.
\newblock In {\em International Conference on Machine Learning}, pages 7105--7114. PMLR, 2019.

\bibitem{yuan2020plug}
Xin Yuan, Yang Liu, Jinli Suo, and Qionghai Dai.
\newblock Plug-and-play algorithms for large-scale snapshot compressive imaging.
\newblock In {\em Proceedings of the IEEE/CVF Conference on Computer Vision and Pattern Recognition}, pages 1447--1457, 2020.

\bibitem{yuan2021plug}
Xin Yuan, Yang Liu, Jinli Suo, Fr{\'e}do Durand, and Qionghai Dai.
\newblock Plug-and-play algorithms for video snapshot compressive imaging.
\newblock {\em IEEE Transactions on Pattern Analysis and Machine Intelligence}, 44(10):7093--7111, 2021.

\bibitem{zela2022surrogate}
Arber Zela, Julien~Niklas Siems, Lucas Zimmer, Jovita Lukasik, Margret Keuper, and Frank Hutter.
\newblock Surrogate nas benchmarks: Going beyond the limited search spaces of tabular nas benchmarks.
\newblock In {\em Tenth International Conference on Learning Representations}, pages 1--36. OpenReview. net, 2022.

\bibitem{zhang2022herosnet}
Xuanyu Zhang, Yongbing Zhang, Ruiqin Xiong, Qilin Sun, and Jian Zhang.
\newblock Herosnet: Hyperspectral explicable reconstruction and optimal sampling deep network for snapshot compressive imaging.
\newblock In {\em Proceedings of the IEEE/CVF Conference on Computer Vision and Pattern Recognition}, pages 17532--17541, 2022.

\bibitem{zhao2016spectral}
Wenzhi Zhao and Shihong Du.
\newblock Spectral--spatial feature extraction for hyperspectral image classification: A dimension reduction and deep learning approach.
\newblock {\em IEEE Transactions on Geoscience and Remote Sensing}, 54(8):4544--4554, 2016.

\bibitem{zhong2017spectral}
Zilong Zhong, Jonathan Li, Zhiming Luo, and Michael Chapman.
\newblock Spectral--spatial residual network for hyperspectral image classification: A 3-d deep learning framework.
\newblock {\em IEEE Transactions on Geoscience and Remote Sensing}, 56(2):847--858, 2017.

\end{thebibliography}
}

\end{document}